\documentclass{article}

\usepackage{amsmath,amsthm}
\usepackage[dvipsnames,table]{xcolor}

\usepackage[parfill]{parskip}
\usepackage[citestyle=numeric-comp,sorting=nyt,maxbibnames=99,backend=biber]{biblatex}
\newcommand{\citep}{\parencite}
\newcommand{\citet}{\textcite}
\usepackage[T1]{fontenc}
\usepackage[tt=false, type1=true]{libertine}
\usepackage[varqu]{zi4}
\usepackage[libertine]{newtxmath}

\usepackage{bm}
\usepackage{mathtools}
\usepackage{nicefrac}

\usepackage{booktabs}
\usepackage{microtype}
\usepackage{multirow}
\usepackage[inline]{enumitem}
\usepackage{graphicx}
\usepackage[font=small]{caption}
\usepackage{float}
\usepackage{pgfplots}
\pgfplotsset{compat=1.18}
\usetikzlibrary{patterns}

\usepackage{url}
\usepackage{hyperref}
\definecolor{link}{HTML}{307ef1}
\hypersetup{
     colorlinks   = true,
     linkcolor    = link,
     citecolor    = link,
     urlcolor     = link,
     pdfborderstyle={/S/U/W 1},
}

\usepackage[capitalise,noabbrev]{cleveref}

\usepackage{xspace}

\title{Distilling Token-Trained Models into Byte-Level Models}
\author{%
  Zishuo Bao$^{1,2}$ \quad Jiaqi Leng$^{3}$ \quad Junxiong Wang$^{4}$ \quad Bowen Peng$^{5}$ \quad Yucheng Lu$^{1,2}$ \\[0.6em]
  $^{1}$New York University \quad $^{2}$NYU Shanghai \quad $^{3}$Fudan University \quad $^{4}$Together AI \quad $^{5}$Nous Research \\[0.3em]
  \texttt{\{zb2466, lu.yucheng\}@nyu.edu}
}
\date{}

\begin{document}

\maketitle

\begin{abstract}
Byte Language Models (BLMs) have emerged as a promising direction for scaling language models beyond tokenization. However, existing BLMs typically require training from scratch on trillions of bytes, making them prohibitively expensive. In this paper, we propose an efficient distillation recipe that converts existing token-trained LLMs into BLMs while retaining comparable capabilities. Our recipe follows a two-stage curriculum: (1) \emph{Progressive Knowledge Distillation}, which aligns byte-level representations with the embeddings of the token-trained teacher model; and (2) \emph{Byte-Level Supervised Fine-Tuning}, which enables end-to-end generation entirely in the byte space.
We validate our approach across multiple model families, including Llama, Qwen, and OLMo, and demonstrate that the distilled BLMs retain most of the teacher models’ performance using only approximately 125B bytes.
\end{abstract}

\section{Introduction}
Recent developments have witnessed a surge of interest in Byte Language Models (BLMs), which operate directly in the byte space by consuming and generating bytes, thereby eliminating the inductive bias imposed by tokenization. A growing body of work has demonstrated that modeling language at the byte level can achieve strong performance while mitigating tokenization-induced artifacts \citep{nawrot2022hierarchical,nawrot2023efficient,yu2023megabyte,slagle2024spacebyte,ho2024block,wang2024mambabytetokenfreeselectivestate,pagnoni2024bytelatenttransformerpatches,hwang2025dynamicchunkingendtoendhierarchical}.

A common thread across these successful architectures is the use of a \emph{hierarchical structure}: they typically leverage a lightweight encoder to first compress raw bytes into chunks, which are then processed by a transformer backbone before being decoded back into bytes. This structural similarity to standard Large Language Models (LLMs) suggests a natural opportunity for knowledge transfer. In particular, it enables initializing the transformer backbone from powerful pretrained token-based (subword) LMs.

However, while several prior works explore initializing byte-level models with weights from pretrained token-trained LMs to accelerate convergence \citep{hwang2025dynamicchunkingendtoendhierarchical,pagnoni2024bytelatenttransformerpatches}, these approaches have not succeeded in closing the performance gap: the resulting BLMs still substantially underperform their token-based counterparts.
To address this gap, in this paper we study the following question: \emph{Can we obtain a byte-level model through distillation from a token-trained model, without training from scratch?}

\begin{figure}[t!]
    \centering
    \includegraphics[width=\linewidth]{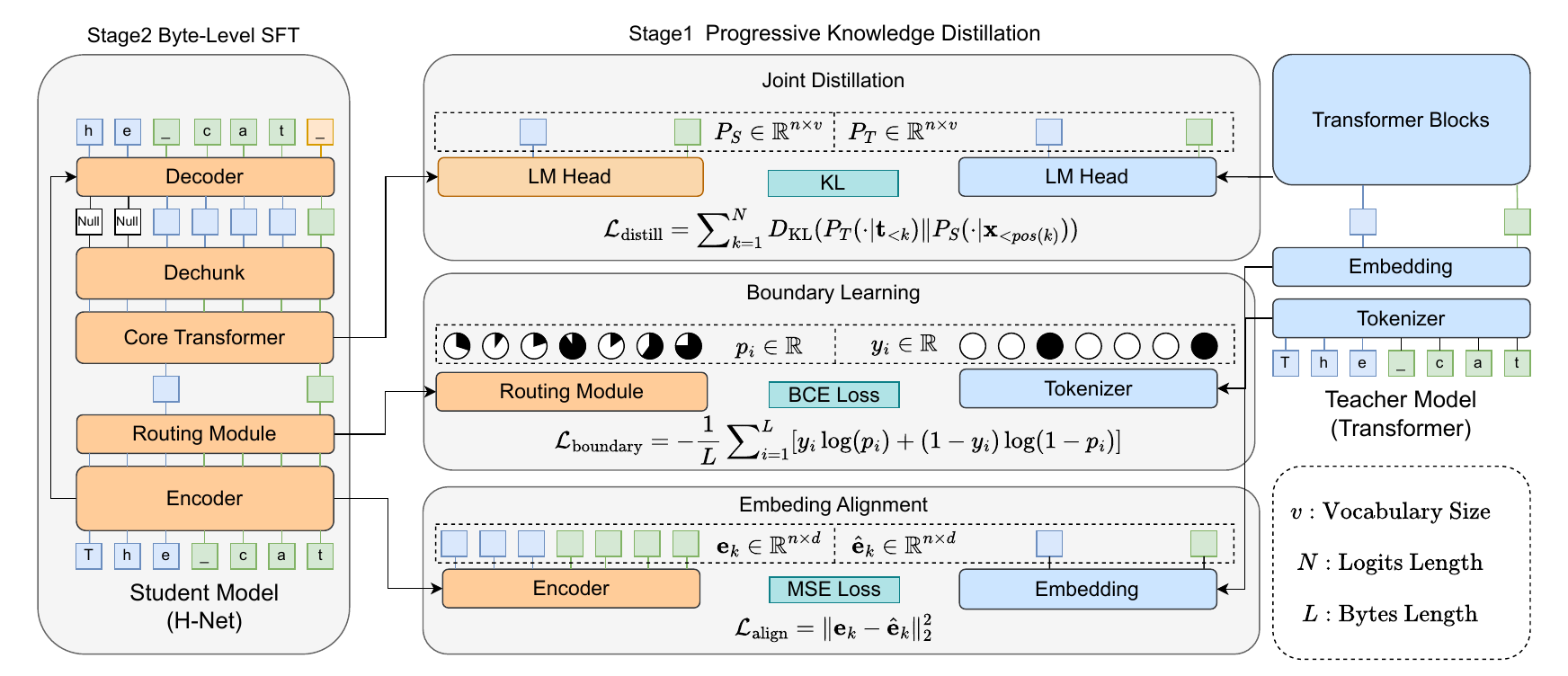}
    \caption{Overview of the Two-Stage Distillation Framework.
    \textbf{Stage 1: Progressive Knowledge Distillation} consists of three steps:
    (a) \textit{Embedding Alignment} to map byte patches to the token space ($L_{align}$); 
    (b) \textit{Joint Distillation} to synchronize hidden states and logits ($L_{distill}$); 
    (c) \textit{Boundary Learning} to align tokenization via the One-Byte Lookahead Router ($L_{boundary}$). 
    \textbf{Stage 2: Full Byte-Level SFT} enables end-to-end generation in the byte space via Dechunk and Decoder modules. }
    \label{fig:overview}
\end{figure}

We answer this question in the affirmative by proposing a novel distillation recipe. Our approach distills a standard Transformer-based language model into a hierarchical byte-level model, such as H-Net \citep{hwang2025dynamicchunkingendtoendhierarchical}.
The central challenge here lies in the boundary mismatch between the two representations: token-based models impose fixed, discrete boundaries over raw byte sequences, whereas byte models dynamically group bytes into variable-length units. This mismatch falls outside the scope of traditional distillation methods, which typically assume a fixed and shared output space.
To resolve this, as illustrated in Figure~\ref{fig:overview}, we introduce a two-stage distillation pipeline that progressively aligns fine-grained byte-level representations with pretrained token embeddings. Unlike prior approaches that rely on extensive retraining from scratch, our method follows a structured curriculum:

\textbf{Stage 1: Progressive Knowledge Distillation.} The main challenge in distillation arises from mismatched boundaries between LLMs and BLMs in byte space. We address this challenge with a three-step process: (1) embedding alignment, (2) joint distillation, and (3) boundary learning, which trains the student model to process raw bytes while faithfully mimicking the teacher model’s token-level outputs. This stage allows the distilled BLM to take raw bytes as inputs and generate in the token space.

\textbf{Stage 2: Byte-Level Supervised Fine-Tuning (SFT).} Building on Stage~1, we introduce a dechunking module and apply supervised fine-tuning (SFT) with a next-byte prediction objective, enabling the model to generate entirely in the byte space.

We substantiate our recipe over a variety of model families, including Llama~\citep{grattafiori2024llama3herdmodels}, Qwen~\citep{yang2025qwen3technicalreport} and OLMo~\citep{olmo20242}.
The empirical results demonstrate unprecedented data efficiency. With approximately 125 B training bytes in total, our distilled model retains the vast majority of teacher models’ capabilities in downstream tasks. For instance, on the MMLU benchmark, our method achieves scores of 51.8 and 68.5 for Llama-3.2 3B and Qwen-3 4B respectively, retaining over 92\% of the original performance. This represents a significant reduction in resource consumption compared with state-of-the-art baselines.

Our main contributions in this paper can be summarized as follows:
\begin{itemize}[nosep,leftmargin=12pt]
    \item We propose a novel two-stage distillation framework that effectively transforms token-trained LLMs into BLMs, which drastically reduces the data requirement compared with training BLMs from scratch.
    \item We validate our approach across various models (including Llama, Qwen, and OLMo). Furthermore, we explore post-training strategies, demonstrating that \textit{On-Policy Distillation} can further enhance the student model's capabilities.
    \item We validate the effectiveness of our architectural choices through extensive ablation. Furthermore, we provide a comprehensive analysis of the intrinsic characteristics and behaviors of the distilled student models on byte-level tasks.
    \item We open-source all components of our recipe, including the distillation code and pretrained model checkpoints, for the community to build upon\footnote{\url{https://github.com/heavyball-research/DistillBytes}}.
\end{itemize}

\section{Preliminaries and Related Work}
\label{RelatedWork}

\subsection{Efficient Byte-Level Architectures}

Directly modeling raw bytes removes tokenization biases but introduces compute challenges from the quadratic cost of self-attention over long sequences. To mitigate this, recent architectures fall into \textit{isotropic} and \textit{hierarchical} approaches.

\textit{Isotropic architectures} process the full byte sequence without downsampling, typically using linear-time mechanisms to bypass the cost of standard attention. For instance, \citet{wang2024mambabytetokenfreeselectivestate} leverages Selective State Space Models (SSMs) for linear efficiency. \textit{Hierarchical architectures} reduce effective sequence length by grouping bytes into coarser chunks, enabling standard transformer backbones. \textit{Static chunking} \citep{nawrot2022hierarchical,yu2023megabyte,ho2024block,egli2025multiscale} partitions inputs into fixed-size chunks; though efficient, content-agnostic boundaries often break semantic units. \textit{External chunking} defines boundaries via heuristics or auxiliary signals: \citet{slagle2024spacebyte} segments by whitespace, while \citet{nawrot2023efficient} and \citet{pagnoni2024bytelatenttransformerpatches} use entropy-based detection of information-dense transitions. \textit{Dynamic chunking} learns segmentation end-to-end within the model: \citet{hwang2025dynamicchunkingendtoendhierarchical} uses a learnable routing module to segment by intrinsic semantic granularity. We adopt H-Net as our backbone for its balance between efficiency and linguistic plausibility.

\subsection{Token-to-Byte Knowledge Distillation}
Knowledge distillation (KD)~\citep{hinton2015distillingknowledgeneuralnetwork} is a fundamental technique for transferring capabilities from teacher to student. Prior work explores cross-architecture distillation~\citep{kasai2021finetuning,wang2024mamba} from Transformers~\citep{vaswani2017attention} to linear RNNs~\citep{katharopoulos2020transformers,dao2024transformersssmsgeneralizedmodels} or hybrids. While KD transfers knowledge across architectures with a shared tokenizer, token-to-byte distillation faces unique challenges due to boundary misalignment from varying vocabularies and sequence lengths \citep{minixhofer2025universalcrosstokenizerdistillationapproximate}.

In byte-level modeling, \citet{pagnoni2024bytelatenttransformerpatches} initializes latent transformers from pretrained token models (e.g., Llama 3.1 8B), and \citet{hwang2025dynamicchunkingendtoendhierarchical} starts from token LM weights with auxiliary alignment losses. However, these are essentially SFT and still demand large data to replicate teacher capabilities.

Recently, Bolmo introduced a \emph{byteification} process that employs a direct distillation strategy \citep{minixhofer2025bolmobyteifyinggenerationlanguage}, which makes the first attempt with multiple approximation methods for vocabulary and sequence alignment. 
Although these approximations work well for OLMo families~\citep{olmo20242, olmo2025olmo}, their applicability to other model families remains uncertain, likely necessitating substantial manual hyperparameter tuning across different settings.

\section{Methodology}
In this section, we introduce our main distillation recipe. We build our student model on H-Net~\citep{hwang2025dynamicchunkingendtoendhierarchical}, a representative hierarchical byte-level architecture, and adapt it for distillation from a token-trained teacher. Section~\ref{sec:architecture_design} describes the architectural modifications, and Section~\ref{sec:distillation_framework} presents our two-stage distillation framework.

\subsection{Student Model Architecture}
\label{sec:architecture_design}

\paragraph{Background: H-Net.} As shown on the left of Figure~\ref{fig:overview}, H-Net is a U-Net-like hierarchical model that processes raw bytes through five components arranged from input to output: an \emph{Encoder} that maps the byte sequence into latent representations; a \emph{Routing Module} that predicts a boundary probability for each byte, and dynamically segments the sequence into variable-length chunks; a \emph{Core Transformer} that operates on the resulting compressed chunk sequence; a \emph{Dechunk} module that upsamples chunk representations back to byte-level resolution; and a \emph{Decoder} that produces the final byte-level outputs. We adopt H-Net's overall structure, and initialize the Core Transformer with the teacher model's weights, exploiting their structural compatibility with standard transformer LMs.

To make H-Net suitable for distillation from a token-trained teacher, three modifications are required: aligning chunk boundaries with token endings (Section~\ref{sec:lookahead}), enabling causal autoregressive generation (Section~\ref{sec:decoding}), and preserving causality during chunk-to-byte upsampling (Section~\ref{sec:causality}).

\subsubsection{One-Byte Lookahead Routing}
\label{sec:lookahead}

H-Net's original Routing Module compares each byte's hidden state with its \emph{predecessor}, treating high-dissimilarity positions as \emph{chunk starts}. However, subword tokenizers conventionally place boundaries at \emph{chunk endings}. To align with this convention, we shift the comparison forward by one position, computing the boundary probability between the current and \emph{next} byte:
\begin{equation}
    p = \frac{1}{2}\left(1 - \frac{\mathbf{x}_{\text{curr}}^\top \mathbf{x}_{\text{next}}}{\| \mathbf{x}_{\text{curr}} \| \| \mathbf{x}_{\text{next}} \|}\right),
    \qquad b = \mathbb{I}(p \geq 0.5).
\end{equation}
The indicator $b = 1$ now marks the \emph{end} of a chunk. This \emph{one-byte lookahead} ensures that the hidden state at each chunk boundary aggregates information through the closure of its chunk before entering the Core Transformer.

\subsubsection{Decoding Strategy}
\label{sec:decoding}

The one-byte lookahead routing described above operates naturally during parallel prefill, where all bytes are available. During autoregressive generation, however, the next byte $\mathbf{x}_{\text{next}}$ does not yet exist when the router needs to decide whether the current byte ends a chunk. We explore two strategies to reconcile lookahead with causal decoding.

\textbf{Joint Boundary Prediction (JBP).} A straightforward approach, introduced by \citet{minixhofer2025bolmobyteifyinggenerationlanguage}, sidesteps the routing module at inference time by folding the boundary decision into the output vocabulary. The decoder's vocabulary is expanded from $\mathcal{V}_{\text{byte}}$ to $\mathcal{V}_{\text{byte}} \times \{0, 1\}$, where each byte is paired with a binary boundary indicator. At each step, the model predicts a single distribution over these $512$ joint classes, simultaneously emitting the next byte and declaring whether it terminates a chunk. Training proceeds end-to-end in this augmented space.

\textbf{Multi-Byte Prediction (MBP).} Alternatively, we propose Multi-Byte Prediction, which preserves the original routing logic by explicitly synthesizing the future byte that the router requires. Alongside the primary head that predicts $\mathbf{x}_{t+1}$, we attach an auxiliary head trained to predict $\mathbf{x}_{t+2}$. The predicted embedding of $\mathbf{x}_{t+2}$ then serves as a proxy for $\mathbf{x}_{\text{next}}$ in the routing computation for $\mathbf{x}_{t+1}$, allowing the model to evaluate boundary probabilities under the same lookahead formulation used during training without ever observing future bytes.

The two strategies offer different trade-offs: JBP couples byte and boundary prediction tightly but enlarges the output space, while MBP keeps the original routing intact at the cost of an additional head. We compare them empirically in Section~\ref{sec:decoding_ablation}.

\subsubsection{Causality with Shifted Up-sampling}
\label{sec:causality}

The Dechunk module upsamples the $N$ chunk representations produced by the Core Transformer back to the original byte length $L$, before passing them to the Decoder. A naïve implementation would broadcast each chunk representation $h_k$ to all bytes within its chunk. However, since $h_k$ aggregates information from the \emph{entire} chunk, this leaks future bytes into earlier positions and breaks autoregressive causality~\citep{pagnoni2024bytelatenttransformerpatches}.

Following the dynamic pooling strategy of~\citet{nawrot2023efficient}, we resolve this by \emph{shifting} the up-sampling: $h_k$ is revealed only at the \emph{last} byte of the $k$-th chunk, while all other bytes within the chunk receive $h_{k-1}$ instead (a learned \texttt{NULL} state for the first chunk). Each byte position thus only sees chunk representations whose contents have already been emitted, preserving causality while still providing maximal context at chunk boundaries where the next-byte prediction is most informative.

\subsection{Two-stage Distillation Framework}
\label{sec:distillation_framework}

The architectural modifications in Section~\ref{sec:architecture_design} make H-Net structurally compatible with a token-trained teacher, but they do not by themselves transfer the teacher's capabilities. The central difficulty is that the teacher operates over discrete tokens with fixed boundaries, while the student must learn to derive comparable representations from raw bytes through its own dynamic chunking. Closing this gap with a single end-to-end objective is unstable in practice, as we show in Appendix~\ref{appendix:e2e_vs_sequential}: the boundary loss and the distillation loss compete for the encoder's representational capacity, and the student degrades on every benchmark, most severely on tasks that depend on stable sequence-level representations (e.g., HellaSwag, PIQA, and ARC).

We therefore decompose distillation into two stages (Figure~\ref{fig:overview}): \textbf{Stage 1 (Progressive Knowledge Distillation)} aligns byte-level representations with the teacher's token-level distribution while retaining its LM head, and \textbf{Stage 2 (Byte-Level SFT)} replaces the LM head with the Dechunk and Decoder modules to transition generation into the byte space.

\subsubsection{Stage 1: Progressive Knowledge Distillation}
The main challenges in the distillation process are the \emph{vocabulary} and \emph{sequence length} mismatch: while the teacher model operates on discrete semantic tokens, the student model must learn to derive these chunks from raw byte sequences. To address this, we design three specific loss objectives to align the student model's embedding space, semantic probabilities, and boundary decisions with those of the teacher model, which are illustrated in the middle of Figure \ref{fig:overview}.

\textbf{Loss Definitions.}
First, to handle mismatches, we define the \textit{Embedding Alignment Objective} ($\mathcal{L}_{\text{align}}$). For a byte span corresponding to a token $t_k$, we minimize the difference between the hidden state of the student model encoder at the boundary byte $\hat{\mathbf{e}}_k$ and the static embedding of the teacher model $\mathbf{e}_k$:
\begin{equation}
\mathcal{L}_{\text{align}} = \frac{1}{N} \sum_{k=1}^{N} \|\mathbf{e}_k - \hat{\mathbf{e}}_k\|_2^2
\end{equation}
where $N$ denotes the total number of tokens in the sequence.

Second, to further transfer capabilities of teacher model, we employ the \textit{Teacher-Forced Distillation Objective} ($\mathcal{L}_{\text{distill}}$). By utilizing the teacher model's segmentation boundaries to synchronize the sequence lengths, we minimize the KL divergence between the student model's and teacher model's output distributions:
\begin{equation}
\mathcal{L}_{\text{distill}} = \sum_{k=1}^{N} D_{\text{KL}}(P_T(\cdot | \mathbf{t}_{<k}) \| P_S(\cdot | \mathbf{x}_{\le \mathrm{pos}(k-1)}))
\end{equation}
Here, $\mathbf{t}_{<k}$ represents the teacher model's prefix tokens, $\mathrm{pos}(k)$ is the index of the last byte of the $k$-th token (with $\mathrm{pos}(0) = 0$), and $\mathbf{x}_{\le \mathrm{pos}(k-1)}$ is the student model's byte prefix that ends at the boundary of the $(k-1)$-th token, i.e., the same context as $\mathbf{t}_{<k}$.

Third, to enable autonomous tokenization, we introduce the \textit{Boundary Learning Objective} ($\mathcal{L}_{\text{boundary}}$), where the Routing Module is trained to predict token boundaries via binary classification on each byte:
\begin{equation}
\mathcal{L}_{\text{boundary}} = - \frac{1}{L} \sum_{i=1}^{L} [y_i \log(p_i) + (1-y_i) \log(1-p_i)]
\end{equation}
where $L$ is the total length of the byte sequence, $y_i \in \{0, 1\}$ is the ground-truth boundary label derived from the teacher tokenizer, and $p_i$ is the predicted probability that byte $i$ is a boundary.

\textbf{Optimization Strategy.}
While a holistic objective $\mathcal{L}_{\text{total}} = \lambda_{1}\mathcal{L}_{\text{align}} + \lambda_{2}\mathcal{L}_{\text{distill}} + \lambda_{3}\mathcal{L}_{\text{boundary}}$, as adopted in \citep{minixhofer2025bolmobyteifyinggenerationlanguage}, is theoretically possible, it often leads to training instability due to the competition between boundary learning and semantic alignment and requires extensive hand-tuning of multiple hyperparameters. 

In contrast, our framework adopts a \textit{sequential curriculum} that eliminates these sensitivities:
\begin{enumerate}[nosep, leftmargin=12pt]
    \item \textbf{Step 1 (Embedding Alignment):} Optimize $\mathcal{L}_{\text{align}}$ to establish a foundational mapping from bytes to the teacher model's space.
    \item \textbf{Step 2 (Joint Distillation):} Train the Encoder and Core Transformer using $\mathcal{L}_{\text{distill}}$ to adapt the backbone to byte-derived representations.
    \item \textbf{Step 3 (Boundary Learning):} Freeze the backbone and optimize $\mathcal{L}_{\text{boundary}}$ to finalize the Routing Module.
\end{enumerate}
This curriculum ensures that each module is trained on top of a stable representation, making the framework highly robust and generalizable to diverse models without extensive hyperparameter searching.

Upon completion of this stage, the student model successfully learns to dynamically group raw bytes into chunks and generate in the token space.

\subsubsection{Stage 2: SFT for Byte-to-Byte Modeling}
\label{sec:sft}
The student model after Stage 1 still relies on a token LM head for generation, necessitating the subsequent stage for end-to-end byte generation. We replace the token LM head with \textit{Dechunk} and \textit{Decoder} modules and train the model using the Next Byte Prediction loss.

Stage 2 consists of two fine-tuning steps. First, we perform \textit{Head Adaptation}, training only the newly initialized Dechunk and Decoder modules to align with the distilled backbone representations. Subsequently, we proceed to \textit{end-to-end fine-tuning}, where all parameters are unfrozen to optimize the entire model. This ensures that the model retains the linguistic richness acquired in Stage 1 while gaining the flexibility of pure byte-level modeling.

\begin{table}[t]
\centering
\caption{\textbf{Main Results: Comparison of Byte-Level Distillation Methods.} We evaluate our two-stage recipe against leading baselines across various model families. \textbf{Avg Drop} is the mean, over the benchmarks reported for each method, of the accuracy gap between a student and its own token-trained teacher. For our Stage 2 students, \textbf{Bytes} is the total over both stages (30B in Stage 1 + 95B in Stage 2). Compared to existing baselines, our approach achieves the same performance retention using fewer training bytes.}
\label{tab:main_results}
\renewcommand{\arraystretch}{1.15}
\setlength{\tabcolsep}{3pt}
\resizebox{\textwidth}{!}{%
\begin{tabular}{l|l|c|cccccccc|c}
    \toprule
    \multirow{2}{*}{\textbf{Method}} & \multirow{2}{*}{\textbf{Role / Model}} & \multirow{2}{*}{\textbf{Bytes}} & \multicolumn{8}{c|}{\textbf{Task Performance}} & \textbf{Retention} \\
     & & & \textbf{LMB} & \textbf{HellA} & \textbf{PIQA} & \textbf{ARC-E} & \textbf{ARC-C} & \textbf{WINO} & \textbf{Open} & \textbf{MMLU} & \textbf{Avg Drop} $\downarrow$ \\
    \midrule
    
    \multicolumn{12}{c}{\cellcolor{gray!10}\textit{Existing Leading Baselines}} \\
    \midrule
    \multirow{2}{*}{\shortstack[l]{BLT Distill\\\citep{pagnoni2024bytelatenttransformerpatches}}} 
     & Teacher (Llama 3.1 8B) & - & - & 80.7 & 80.7 & 83.4 & 55.2 & - & - & 66.3 & - \\
     & Student (Byte) & 220B & - & 76.1 & 77.4 & 66.6 & 45.8 & - & - & 63.7 & 7.34 \\
    \midrule
    \multirow{2}{*}{\shortstack[l]{ALM\\\citep{minixhofer2025universalcrosstokenizerdistillationapproximate}}} 
     & Teacher (Llama 3.2 3B) & - & - & - & 76.9 & - & 43.9 & - & - & 62.4 & - \\
     & Student (Byte) & - & - & - & 73.7 & - & 40.1 & - & - & 55.9 & 4.50 \\
    \midrule
    \multirow{2}{*}{\shortstack[l]{H-Net Distill\\\citep{hwang2025dynamicchunkingendtoendhierarchical}}} 
     & Teacher (Llama 3.2 3B) & - & 70.1 & 73.7 & 76.8 & 74.5 & 46.0 & 68.8 & 42.8 & 56.1 & - \\
     & Student (Byte) & 189B & 63.4 & 70.2 & 76.1 & 72.1 & 43.3 & 66.5 & 41.4 & 51.9 & 2.99 \\
    \midrule
    
    \multicolumn{12}{c}{\cellcolor{gray!10}\textit{Ours: Two-stage Distillation Pipeline}} \\
    \midrule
    \multirow{3}{*}{\shortstack[l]{\textbf{Ours}\\(Llama 3.2 3B)}} 
     & \textbf{Teacher (Baseline)} & - & 70.1 & 73.7 & 76.8 & 74.6 & 46.0 & 67.4 & 41.4 & 56.0 & - \\
     & \textbf{Ours (Stage 1)} & 30B & 69.5 & 72.9 & 76.1 & 73.6 & 43.9 & 67.2 & 41.1 & 51.6 & \textbf{1.26} \\
     & \textbf{Ours (Stage 2)} & 125B (30B + 95B) & 69.3 & 71.2 & 73.7 & 69.5 & 41.9 & 65.9 & 39.4 & 51.8 & \underline{2.91} \\
    \midrule
    \multirow{3}{*}{\shortstack[l]{\textbf{Ours}\\(Qwen 3 4B)}} 
     & \textbf{Teacher (Baseline)} & - & 69.1 & 73.6 & 77.7 & 78.9 & 51.3 & 70.4 & 40.6 & 73.0 & - \\
     & \textbf{Ours (Stage 1)} & 30B & 68.4 & 72.5 & 77.0 & 76.8 & 50.2 & 69.0 & 40.2 & 66.8 & \textbf{1.71} \\
     & \textbf{Ours (Stage 2)} & 125B (30B + 95B) & 66.2 & 69.7 & 75.4 & 77.1 & 52.6 & 66.4 & 41.2 & 68.5 & \underline{2.19} \\
    \bottomrule
\end{tabular}%
}
\end{table}

\begin{table}[t]
\centering
\caption{\textbf{Impact of On-Policy Distillation in Stage 1.} We compare the Llama-3.2 3B Instruct teacher model with its byte-level student model before and after on-policy distillation. On-policy distillation improves performance across nearly all benchmarks and narrowing the gap with the teacher.}
\label{tab:post_training}
\small
\setlength{\tabcolsep}{4pt}
\renewcommand{\arraystretch}{0.95}
\begin{tabular}{l|c|cccccccc}
\toprule
\textbf{Model Phase} & \textbf{Role} & \textbf{LMB} & \textbf{HellA} & \textbf{PIQA} & \textbf{ARC-E} & \textbf{ARC-C} & \textbf{WINO} & \textbf{Open} & \textbf{MMLU} \\
\midrule
\textbf{Baseline} & Teacher & 65.98 & 70.40 & 75.79 & 74.03 & 45.82 & 67.64 & 36.00 & 59.68 \\
Stage 1 & Student & 49.16 & 69.09 & 75.24 & 72.01 & 43.52 & 65.98 & \textbf{36.20} & 52.52 \\
\textbf{+ On-Policy} & Student & \textbf{55.95} & \textbf{69.44} & \textbf{76.06} & \textbf{72.52} & \textbf{45.05} & \textbf{66.93} & 36.00 & \textbf{54.64} \\
\bottomrule
\end{tabular}
\end{table}

\subsubsection{Comparison with Bolmo}
Our framework diverges from Bolmo \citep{minixhofer2025bolmobyteifyinggenerationlanguage} in two respects. First, while Bolmo performs distillation in a single end-to-end step, we decompose it into a sequential curriculum that separates representation alignment and capability transfer from boundary learning: the model first aligns encoder representations and transfers capabilities from the teacher, then the backbone is frozen to learn dynamic byte boundaries. Second, the paradigms differ: Bolmo trains the student to directly approximate the target byte-level distribution, whereas we introduce an intermediate stage in which the student learns to dynamically group raw bytes while generating in the token space. By retaining the teacher's token-level LM head during Stage 1, we align the student with the teacher's token-level distribution before transitioning to fully byte-level generation. A detailed empirical comparison against Bolmo 1B and other baselines is provided in Appendix \ref{app:bolmo_comparison}.

\section{Experiments}
\label{sec:experiments}

Our experiments aim to verify the data efficiency of our two-stage curriculum and the plasticity of the model for post-training. Following the setup in Section \ref{sec:setup}, Section \ref{sec:performance} compares our framework with baselines such as BLT and H-Net, highlighting the critical role of the intermediate alignment stage. Finally, Section \ref{sec:post_training} validates the model's extensibility through standard on-policy optimization.

\subsection{Experimental Setup}
\label{sec:setup}

\textbf{Models and Architecture.} To evaluate the universality of the proposed approach, experiments are conducted on two open-weight model families: Llama-3.2 3B and Qwen-3 4B. These models function as token-trained LM teacher models. In the student models, the core transformer layers and language modeling head are initialized with the corresponding teacher model's weights. We further apply the recipe to OLMo~2 1B~\citep{olmo20242}; these results are reported in Appendix~\ref{appendix:e2e_vs_sequential} and Appendix~\ref{app:bolmo_comparison}. Notice that all the models employ the JBP decoding strategy described in Section~\ref{sec:decoding}. A detailed ablation study of the two decoding strategies is provided in Section~\ref{sec:decoding_ablation}. The detailed model configurations are provided in Appendix \ref{appendix:hyperparameters}.

\textbf{Training Data.} We utilize the FineWeb dataset~\citep{penedo2024the} for all training phases. Stage 1 is conducted on a 30B byte subset (approximately 7B tokens), which provides sufficient semantic coverage for knowledge transfer. For Stage 2, we further employ 95B bytes of data to refine the model's autonomous generation capabilities. In total, our framework requires only 125B bytes of training data.

\textbf{Evaluation Metrics.} The performance is evaluated on a comprehensive suite of downstream benchmarks: LAMBADA (LMB)~\citep{paperno2016lambada}, HellaSwag (HellA)~\citep{zellers2019hellaswagmachinereallyfinish}, PIQA~\citep{bisk2019piqareasoningphysicalcommonsense}, ARC-Easy (ARC-E), ARC-Challenge (ARC-C)~\citep{clark2018thinksolvedquestionanswering}, WinoGrande (WINO)~\citep{sakaguchi2019winograndeadversarialwinogradschema}, OpenBookQA (Open)~\citep{mihaylov2018suitarmorconductelectricity}, and Massive Multitask Language Understanding (MMLU)~\citep{hendrycks2021measuringmassivemultitasklanguage}. We additionally report GSM8K~\citep{cobbe2021training} in the decoding-strategy ablation (Section~\ref{sec:decoding_ablation}) and the teacher-scale study (Appendix~\ref{appendix:teacher_scale}). For the OLMo~2 1B experiments (Appendix~\ref{appendix:e2e_vs_sequential} and Appendix~\ref{app:bolmo_comparison}), we report 5-shot accuracy on ARC (the average of ARC-E and ARC-C), MMLU, CommonsenseQA (CSQA)~\citep{talmor2019commonsenseqa}, HellaSwag, WinoGrande, SocialIQA~\citep{sap2019socialiqa}, and PIQA. We do all of the evaluations using \textit{lm-evaluation-harness}~\citep{eval-harness} for reproducibility.

\textbf{Hyperparameters.}
Across all training stages, we utilize the AdamW optimizer with a cosine decay schedule and a weight decay of $0.1$. The detailed hyperparameter configurations for each stage are provided in Appendix \ref{appendix:hyperparameters}. All experiments are conducted on a single node of 8$\times$NVIDIA H100 GPUs.

\subsection{Effectiveness of the Proposed Recipe}
\label{sec:performance}

\subsubsection{Performance of Stage 1}
\textbf{Comparison with Leading Baselines.} We evaluated our method against leading byte-level distillation baselines: BLT \citep{pagnoni2024bytelatenttransformerpatches}, ALM \citep{minixhofer2025universalcrosstokenizerdistillationapproximate}, and H-Net \citep{hwang2025dynamicchunkingendtoendhierarchical}. Given the differences in base model capabilities, we report \textit{Avg Drop} to isolate distillation efficiency from the varying strengths of teacher models, which is defined as the mean accuracy reduction of the student model relative to its specific teacher model across the reported tasks. In Table~\ref{tab:main_results}, baseline rows are quoted from the respective original papers. For our Llama-3.2 3B teacher, LAMBADA, HellaSwag, PIQA, and ARC-Challenge are quoted from H-Net~\citep{hwang2025dynamicchunkingendtoendhierarchical}; all other entries in our rows are evaluated with lm-evaluation-harness v0.4.9.

As shown in Table \ref{tab:main_results}, our Stage~1 students have the lowest Avg Drop of all methods: 1.26 and 1.71 points for Llama-3.2 3B and Qwen-3 4B, respectively, compared with 7.34 for BLT and 4.50 for ALM. With only 30B bytes, our Llama-3.2 3B Stage~1 student matches or exceeds the 189B-byte H-Net student on 6 of the 8 benchmarks.

\definecolor{myblue}{RGB}{31, 119, 180}
\definecolor{mygreen}{RGB}{44, 160, 44}
\definecolor{myorange}{RGB}{255, 127, 14}
\definecolor{myred}{RGB}{214, 39, 40}

\begin{figure}[h]
    \centering
    \begin{tikzpicture}
        \begin{axis}[
            width=\linewidth, height=5.5cm,
            xlabel={Training Amount (Billions of Bytes)},
            ylabel={Accuracy},
            xmin=1, xmax=20.5,
            ymin=0.2, ymax=0.9,
            xtick={1,3,5,7,9,11,13,15,17,19},
            grid=major,
            grid style={dashed, gray!30},
            legend style={
                at={(0.5,1.02)},
                anchor=south,
                legend columns=-1,
                font=\footnotesize,
                cells={anchor=west},
                /tikz/every even column/.append style={column sep=10pt},
                draw=none, fill=none,
            },
            tick label style={font=\footnotesize},
            label style={font=\small},
        ]

        \draw [gray, thick, densely dotted] (axis cs:10.5, \pgfkeysvalueof{/pgfplots/ymin}) -- (axis cs:10.5, \pgfkeysvalueof{/pgfplots/ymax});
        
        \node[anchor=north, align=center, font=\small] at (axis cs:5.5, 0.9) {\textbf{Step 1:}\\\textbf{Embedding Alignment}};
        \node[anchor=north, align=center, font=\small] at (axis cs:15.5, 0.9) {\textbf{Step 2:}\\\textbf{Joint Distillation}};

        \addplot[
            color=mygreen,
            mark=*,
            mark size=1.5pt,
            thick,
        ] coordinates {
            (1,0.58)(2,0.69)(3,0.725)(4,0.74)(5,0.755)(6,0.755)(7,0.758)(8,0.757)(9,0.756)(10,0.759)
            (11,0.76)(12,0.762)(13,0.765)(14,0.762)(15,0.763)(16,0.760)(17,0.762)(18,0.760)(19,0.758)(20,0.762)
        };
        \addlegendentry{PIQA}

        \addplot[
            color=myblue,
            mark=square*,
            mark size=1.5pt,
            thick,
        ] coordinates {
            (1,0.38)(2,0.62)(3,0.67)(4,0.70)(5,0.705)(6,0.71)(7,0.712)(8,0.708)(9,0.713)(10,0.711)
            (11,0.725)(12,0.73)(13,0.732)(14,0.735)(15,0.735)(16,0.735)(17,0.735)(18,0.735)(19,0.735)(20,0.736)
        };
        \addlegendentry{Hellaswag}

        \addplot[
            color=myorange,
            mark=triangle*,
            mark size=2pt,
            thick,
        ] coordinates {
            (1,0.24)(2,0.36)(3,0.398)(4,0.403)(5,0.415)(6,0.419)(7,0.425)(8,0.423)(9,0.424)(10,0.426)
            (11,0.429)(12,0.428)(13,0.433)(14,0.438)(15,0.438)(16,0.436)(17,0.436)(18,0.433)(19,0.433)(20,0.449)
        };
        \addlegendentry{ARC-C}

        \addplot[
            color=myred,
            mark=diamond*,
            mark size=2pt,
            thick,
        ] coordinates {
            (1,0.25)(2,0.31)(3,0.36)(4,0.38)(5,0.39)(6,0.395)(7,0.40)(8,0.398)(9,0.396)(10,0.398)
            (11,0.481)(12,0.496)(13,0.506)(14,0.511)(15,0.511)(16,0.513)(17,0.512)(18,0.516)(19,0.514)(20,0.516)
        };
        \addlegendentry{MMLU}
        \end{axis}
    \end{tikzpicture}
    \caption{\textbf{Training Dynamics across Stage 1.} We track the performance of the Llama-3.2 3B student over the first 20B bytes of Stage 1: 10B bytes of Step 1 (Embedding Alignment), followed by 10B bytes of Step 2 (Joint Distillation). The dotted line marks the transition between the two steps. Step 3 (Boundary Learning) is not shown. All evaluations use ground-truth token boundaries from the teacher tokenizer, since the Routing Module is trained only in Step 3. While simple tasks (e.g., PIQA) converge rapidly during Embedding Alignment (Step 1), knowledge-intensive tasks like MMLU exhibit a distinct phase transition, requiring the distribution matching of Joint Distillation (Step 2) to unlock performance gains. }
    \label{fig:training_curves}
\end{figure}

\begin{table}[t]
\centering
\caption{\textbf{Necessity of Stage 1 Embedding Alignment.} We conduct an ablation using Llama-3.2 3B to evaluate the impact of embedding alignment before full distillation. Results show that omitting embedding alignment leads to a significant performance collapse across all benchmarks, confirming it as a critical foundation for our recipe.}
\label{tab:ablation_stage1}
\small
\setlength{\tabcolsep}{4pt}
\begin{tabular}{l|ccccccc}
\toprule
\textbf{Method} & \textbf{LMB} & \textbf{HellA} & \textbf{PIQA} & \textbf{ARC-E} & \textbf{ARC-C} & \textbf{WINO} & \textbf{Open} \\
\midrule
Baseline & 70.1 & 73.7 & 76.8 & 74.6 & 46.0 & 67.4 & 41.4 \\
Distill & \textbf{70.7} & \textbf{73.6} & \textbf{76.2} & \textbf{74.3} & \textbf{44.9} & \textbf{69.4} & \textbf{41.2} \\
Distill w/o emb & 31.2 & 46.7 & 58.8 & 37.0 & 21.6 & 60.3 & 33.4 \\
\bottomrule
\end{tabular}
\end{table}

\begin{table}[t]
\caption{\textbf{Comparison of Decoding Strategies.} We evaluate Joint Boundary Prediction (JBP) versus Multi-Byte Prediction (MBP) using the Qwen-based student model. Although MBP shows marginal gains on some common-sense tasks, JBP exhibits superior robustness in complex reasoning, as evidenced by the substantial performance lead in GSM8K.}
\centering
\label{tab:ablation_decoding}
\small
\setlength{\tabcolsep}{4pt}
\begin{tabular}{lccccccccc}
    \toprule
    \textbf{Decoding Strategy} & \textbf{LMB} & \textbf{HellA} & \textbf{PIQA} & \textbf{ARC-E} & \textbf{ARC-C} & \textbf{WINO} & \textbf{Open} & \textbf{MMLU} & \textbf{GSM8K} \\
    \midrule
    JBP & 66.2 & 69.7 & \textbf{75.4} & 77.1 & 52.6 & 66.4 & \textbf{41.2} & 68.5 & \textbf{53.5} \\
    MBP & \textbf{66.7} & \textbf{70.3} & 74.8 & \textbf{77.8} & \textbf{53.4} & \textbf{66.5} & 38.2 & \textbf{69.2} & 37.4 \\
    \bottomrule
    \end{tabular}
\end{table}

\begin{table}[t]
\caption{\textbf{Impact of Encoder Architecture.} We compare the performance of Mamba2 against a standard Transformer encoder on Qwen-3 4B after Step 1 (Embedding Alignment). Both variants are evaluated with ground-truth token boundaries from the teacher tokenizer. Results indicate that the Transformer's attention mechanism struggles to align fine-grained byte sequences with high-level semantic embeddings, leading to a significant performance collapse, particularly on knowledge-intensive benchmarks like MMLU.}
\centering
\label{tab:ablation_encoder}
\small
\setlength{\tabcolsep}{4pt}
\begin{tabular}{l|cccccccc}
    \toprule
    \textbf{Encoder} & \textbf{LMB} & \textbf{HellA} & \textbf{PIQA} & \textbf{ARC-E} & \textbf{ARC-C} & \textbf{WINO} & \textbf{Open} & \textbf{MMLU} \\
    \midrule
    \textbf{Mamba2} & \textbf{67.8} & \textbf{73.5} & \textbf{78.2} & \textbf{77.1} & \textbf{49.7} & \textbf{69.5} & \textbf{40.0} & \textbf{70.2} \\
    Transformer & 45.7 & 68.3 & 73.3 & 69.4 & 44.5 & 66.2 & 37.4 & 31.3 \\
    \bottomrule
    \end{tabular}
\end{table}

\textbf{Analysis of Distillation Dynamics.}
We further plot the training dynamics in Stage 1. Figure \ref{fig:training_curves} presents the evaluation accuracy during \textit{Step 1 (Embedding Alignment)} and \textit{Step 2 (Joint Distillation)}.

\textbf{Step 1 (Rapid Alignment):} The sharp increase to the left of the dotted line demonstrates that aligning the encoder with the teacher model's static embedding space effectively solves the boundary misalignment. This process is sufficient to recover the performance of surface-level tasks (e.g., PIQA) with less than 5 B bytes of data.

\textbf{Step 2 (Reasoning Transfer):} The shift to joint distillation is crucial for complex reasoning. While simple tasks reach a plateau, knowledge-intensive tasks such as MMLU and ARC-Challenge continue to show growth in performance. This confirms that while embedding alignment addresses modality, sequence-level distillation is vital for inheriting deep-semantic reasoning.

\textbf{Data Efficiency.} Table \ref{tab:main_results} shows the relationship between training data volume and accuracy. Our Stage~1 student converges with 30B bytes, a six-fold reduction from H-Net's 189B bytes, while matching or exceeding its downstream performance on 6 of the 8 benchmarks.

\subsubsection{Performance of Stage 2}

We summarize the comparison between our Stage 2 model and baseline results in Table~\ref{tab:main_results}.
Our Llama-3.2 3B student model restricts \textit{Avg Drop} to 2.91, comparable to H-Net Distill (2.99) and lower than BLT Distill (7.34) and ALM (4.50). This indicates effective mitigation of representation collapse in byte-level adaptation. Our recipe reaches this retention with 125B training bytes in total, 34\% fewer than the 189B bytes used by H-Net Distill. The robustness extends across architectures, shown by low degradation (2.19) in our Qwen-3 4B student model, confirming our two-stage distillation pipeline's effectiveness over previous SFT approaches.

\subsection{Applicability of Post-Training Techniques}
\label{sec:post_training}

To further validate the extensibility of our framework, we explore whether the Stage 1 model can be enhanced using standard post-training techniques, particularly on-policy optimization, to address performance gaps and enhance reasoning.

\textbf{Methodology.} We integrate \textit{GOLD}~\citep{patino2025_unlocking_on_policy_distillation_for_any_model_family} with \textit{On-Policy Distillation} \citep{lu2025onpolicydistillation}. Although GOLD facilitates token-to-byte distillation, its theoretical effectiveness is linked to an overlap of the vocabularies of the models. This is where the architectural strength of the Stage 1 student model becomes apparent. We use 77k prompts from \textit{Tulu V3 Mix} \citep{lambert2024tulu3}, unfreeze the entire Stage 1 model, and optimize with AdamW (lr $2 \times 10^{-5}$, cosine decay). The Stage 1 BCE loss is added as an auxiliary objective to stabilize the Routing Module. \textit{Llama-3.2 3B Instruct} serves as the teacher to assess instruction following and reasoning transfer.

\textbf{Results.} As shown in Table \ref{tab:post_training}, GOLD and On-Policy distillation consistently boost performance across benchmarks, especially on reasoning tasks: MMLU rises from 52.52 to 54.64. This indicates the Stage 1 architecture is a solid foundation fully compatible with advanced optimization techniques.

\section{Ablation Study}

\subsection{Necessity of Embedding Alignment}

In Stage 1, our framework employs a sequential optimization strategy comprising three steps: embedding alignment, distillation, and boundary learning. To assess the necessity of the initial \textit{embedding alignment} step, we conduct an ablation study using Llama-3.2 3B as the teacher model.

\textbf{Experimental Setup.} We compare two configurations under a fixed data budget of 20B bytes: (1) \textbf{Distill (Sequential)}: The model underwent 10B bytes of embedding alignment, followed by 10B bytes of distillation. (2) \textbf{Distill w/o emb}: The model skips the alignment step and undergoes 20B bytes of direct distillation. For both settings, the learning rates are consistently set to $1 \times 10^{-3}$ for the encoder and $2 \times 10^{-5}$ for the main transformer. The evaluation results for multiple benchmarks are presented in Table~\ref{tab:ablation_stage1}.

\textbf{Analysis.} Omitting embedding alignment causes catastrophic failure: the \textit{Distill w/o emb} variant's LMB drops from 70.7 to 31.2. The byte-level student and token-trained teacher initially occupy different latent spaces; without \textit{embedding alignment} to bridge them, the student's transformer cannot interpret latent states during distillation. Anchoring the student's embedding space to the teacher's enables effective capability transfer.

\subsection{Ablation Study on Decoding Strategy}
\label{sec:decoding_ablation}

We compare the \emph{JBP} and \emph{MBP} decoding strategies (Section \ref{sec:decoding}) used in Stage 2. As shown in Table \ref{tab:ablation_decoding}, while MBP yields marginal gains on some benchmarks (e.g., +0.7 points on MMLU), it suffers severe degradation on GSM8K~\citep{cobbe2021training} ($53.5\%$ vs. $37.4\%$), indicating that MBP is ineffective for long-horizon autoregressive generation: the auxiliary head and similarity-based routing fail to maintain coherence in complex reasoning chains. We thus adopt JBP as the default.

\subsection{Impact of Encoder Architecture}

A pivotal design choice is using \textit{Mamba2} as the encoder. We ablate by replacing the Mamba2 block with a standard transformer of equivalent size, training both under the same \textit{Step 1 (Embedding Alignment)} protocol on Qwen-3 4B. Because the Routing Module is not trained at this point, both variants are evaluated with ground-truth token boundaries from the teacher tokenizer. Their scores are therefore higher than the Stage 1 results in Table~\ref{tab:main_results}, which rely on the learned Routing Module, and should only be compared with each other.

\textbf{Performance Collapse of Attention.} Table \ref{tab:ablation_encoder} shows Mamba2 achieves high fidelity (70.2 MMLU), while the transformer collapses on knowledge-intensive tasks (31.3 MMLU). Raw byte streams are longer than token sequences, hurting attention's ability to capture semantic structure, whereas Mamba2's State Space Model (SSM) backbone effectively compresses these signals into meaningful latent states.

\section{Conclusion}

We propose a two-stage distillation framework that converts pretrained token-based LLMs into byte-level models while preserving most of their capabilities. Our method is effective across Llama, Qwen, and OLMo, achieving competitive performance with only 125B training bytes and substantially lowering the cost of building scalable BLMs.

\printbibliography

@inproceedings{katharopoulos2020transformers,
  title={Transformers are rnns: Fast autoregressive transformers with linear attention},
  author={Katharopoulos, Angelos and Vyas, Apoorv and Pappas, Nikolaos and Fleuret, Fran{\c{c}}ois},
  booktitle={International conference on machine learning},
  pages={5156--5165},
  year={2020},
  organization={PMLR}
}

@article{vaswani2017attention,
  title={Attention is all you need},
  author={Vaswani, Ashish and Shazeer, Noam and Parmar, Niki and Uszkoreit, Jakob and Jones, Llion and Gomez, Aidan N and Kaiser, {\L}ukasz and Polosukhin, Illia},
  journal={Advances in neural information processing systems},
  volume={30},
  year={2017}
}

@article{kasai2021finetuning,
  title={Finetuning pretrained transformers into rnns},
  author={Kasai, Jungo and Peng, Hao and Zhang, Yizhe and Yogatama, Dani and Ilharco, Gabriel and Pappas, Nikolaos and Mao, Yi and Chen, Weizhu and Smith, Noah A},
  journal={arXiv preprint arXiv:2103.13076},
  year={2021}
}

@article{wang2024mamba,
  title={The mamba in the llama: Distilling and accelerating hybrid models},
  author={Wang, Junxiong and Paliotta, Daniele and May, Avner and Rush, Alexander and Dao, Tri},
  journal={Advances in Neural Information Processing Systems},
  volume={37},
  pages={62432--62457},
  year={2024}
}

@misc{grattafiori2024llama3herdmodels,
      title={The Llama 3 Herd of Models}, 
      author={Aaron Grattafiori and Abhimanyu Dubey and Abhinav Jauhri and others},
      year={2024},
      eprint={2407.21783},
      archivePrefix={arXiv},
      primaryClass={cs.AI},
      url={https://arxiv.org/abs/2407.21783}, 
}

@misc{hwang2025dynamicchunkingendtoendhierarchical,
      title={Dynamic Chunking for End-to-End Hierarchical Sequence Modeling}, 
      author={Sukjun Hwang and Brandon Wang and Albert Gu},
      year={2025},
      eprint={2507.07955},
      archivePrefix={arXiv},
      primaryClass={cs.LG},
      url={https://arxiv.org/abs/2507.07955}, 
}

@misc{pagnoni2024bytelatenttransformerpatches,
      title={Byte Latent Transformer: Patches Scale Better Than Tokens}, 
      author={Artidoro Pagnoni and Ram Pasunuru and Pedro Rodriguez and John Nguyen and Benjamin Muller and Margaret Li and Chunting Zhou and Lili Yu and Jason Weston and Luke Zettlemoyer and Gargi Ghosh and Mike Lewis and Ari Holtzman and Srinivasan Iyer},
      year={2024},
      eprint={2412.09871},
      archivePrefix={arXiv},
      primaryClass={cs.CL},
      url={https://arxiv.org/abs/2412.09871}, 
}

@misc{wang2024mambabytetokenfreeselectivestate,
      title={MambaByte: Token-free Selective State Space Model}, 
      author={Junxiong Wang and Tushaar Gangavarapu and Jing Nathan Yan and Alexander M. Rush},
      year={2024},
      eprint={2401.13660},
      archivePrefix={arXiv},
      primaryClass={cs.CL},
      url={https://arxiv.org/abs/2401.13660}, 
}

@misc{dao2024transformersssmsgeneralizedmodels,
      title={Transformers are SSMs: Generalized Models and Efficient Algorithms Through Structured State Space Duality}, 
      author={Tri Dao and Albert Gu},
      year={2024},
      eprint={2405.21060},
      archivePrefix={arXiv},
      primaryClass={cs.LG},
      url={https://arxiv.org/abs/2405.21060}, 
}

@misc{minixhofer2025universalcrosstokenizerdistillationapproximate,
      title={Universal Cross-Tokenizer Distillation via Approximate Likelihood Matching}, 
      author={Benjamin Minixhofer and Ivan Vulić and Edoardo Maria Ponti},
      year={2025},
      eprint={2503.20083},
      archivePrefix={arXiv},
      primaryClass={cs.CL},
      url={https://arxiv.org/abs/2503.20083}, 
}

@misc{zellers2019hellaswagmachinereallyfinish,
      title={HellaSwag: Can a Machine Really Finish Your Sentence?}, 
      author={Rowan Zellers and Ari Holtzman and Yonatan Bisk and Ali Farhadi and Yejin Choi},
      year={2019},
      eprint={1905.07830},
      archivePrefix={arXiv},
      primaryClass={cs.CL},
      url={https://arxiv.org/abs/1905.07830}, 
}

@misc{bisk2019piqareasoningphysicalcommonsense,
      title={PIQA: Reasoning about Physical Commonsense in Natural Language}, 
      author={Yonatan Bisk and Rowan Zellers and Ronan Le Bras and Jianfeng Gao and Yejin Choi},
      year={2019},
      eprint={1911.11641},
      archivePrefix={arXiv},
      primaryClass={cs.CL},
      url={https://arxiv.org/abs/1911.11641}, 
}

@misc{clark2018thinksolvedquestionanswering,
      title={Think you have Solved Question Answering? Try ARC, the AI2 Reasoning Challenge}, 
      author={Peter Clark and Isaac Cowhey and Oren Etzioni and Tushar Khot and Ashish Sabharwal and Carissa Schoenick and Oyvind Tafjord},
      year={2018},
      eprint={1803.05457},
      archivePrefix={arXiv},
      primaryClass={cs.AI},
      url={https://arxiv.org/abs/1803.05457}, 
}

@misc{sakaguchi2019winograndeadversarialwinogradschema,
      title={WinoGrande: An Adversarial Winograd Schema Challenge at Scale}, 
      author={Keisuke Sakaguchi and Ronan Le Bras and Chandra Bhagavatula and Yejin Choi},
      year={2019},
      eprint={1907.10641},
      archivePrefix={arXiv},
      primaryClass={cs.CL},
      url={https://arxiv.org/abs/1907.10641}, 
}

@misc{mihaylov2018suitarmorconductelectricity,
      title={Can a Suit of Armor Conduct Electricity? A New Dataset for Open Book Question Answering}, 
      author={Todor Mihaylov and Peter Clark and Tushar Khot and Ashish Sabharwal},
      year={2018},
      eprint={1809.02789},
      archivePrefix={arXiv},
      primaryClass={cs.CL},
      url={https://arxiv.org/abs/1809.02789}, 
}

@misc{hendrycks2021measuringmassivemultitasklanguage,
      title={Measuring Massive Multitask Language Understanding}, 
      author={Dan Hendrycks and Collin Burns and Steven Basart and Andy Zou and Mantas Mazeika and Dawn Song and Jacob Steinhardt},
      year={2021},
      eprint={2009.03300},
      archivePrefix={arXiv},
      primaryClass={cs.CY},
      url={https://arxiv.org/abs/2009.03300}, 
}

@misc{hinton2015distillingknowledgeneuralnetwork,
      title={Distilling the Knowledge in a Neural Network}, 
      author={Geoffrey Hinton and Oriol Vinyals and Jeff Dean},
      year={2015},
      eprint={1503.02531},
      archivePrefix={arXiv},
      primaryClass={stat.ML},
      url={https://arxiv.org/abs/1503.02531}, 
}

@inproceedings{
  penedo2024the,
  title={The FineWeb Datasets: Decanting the Web for the Finest Text Data at Scale},
  author={Guilherme Penedo and Hynek Kydl{\'\i}{\v{c}}ek and Loubna Ben allal and Anton Lozhkov and Margaret Mitchell and Colin Raffel and Leandro Von Werra and Thomas Wolf},
  booktitle={The Thirty-eight Conference on Neural Information Processing Systems Datasets and Benchmarks Track},
  year={2024},
  url={https://openreview.net/forum?id=n6SCkn2QaG}
}

@misc{minixhofer2025bolmobyteifyinggenerationlanguage,
      title={Bolmo: Byteifying the Next Generation of Language Models}, 
      author={Benjamin Minixhofer and Tyler Murray and Tomasz Limisiewicz and Anna Korhonen and Luke Zettlemoyer and Noah A. Smith and Edoardo M. Ponti and Luca Soldaini and Valentin Hofmann},
      year={2025},
      eprint={2512.15586},
      archivePrefix={arXiv},
      primaryClass={cs.CL},
      url={https://arxiv.org/abs/2512.15586}, 
}

@misc{patino2025_unlocking_on_policy_distillation_for_any_model_family,
  title={Unlocking On-Policy Distillation for Any Model Family},
  author={Carlos Miguel Patiño and Kashif Rasul and Quentin Gallouédec and Ben Burtenshaw and Sergio Paniego and Vaibhav Srivastav and Thibaud Frere and Ed Beeching and Lewis Tunstall and Leandro von Werra and Thomas Wolf},
  year={2025},
  
}

@article{lu2025onpolicydistillation,
  author = {Kevin Lu and Thinking Machines Lab},
  title = {On-Policy Distillation},
  journal = {Thinking Machines Lab: Connectionism},
  year = {2025},
  note = {https://thinkingmachines.ai/blog/on-policy-distillation},
  doi = {10.64434/tml.20251026},
}

@article{lambert2024tulu3,
  title = {Tülu 3: Pushing Frontiers in Open Language Model Post-Training},
  author = {
    Nathan Lambert and 
    Jacob Morrison and 
    Valentina Pyatkin and 
    Shengyi Huang and 
    Hamish Ivison and 
    Faeze Brahman and 
    Lester James V. Miranda and 
    Alisa Liu and 
    Nouha Dziri and 
    Shane Lyu and 
    Yuling Gu and 
    Saumya Malik and 
    Victoria Graf and 
    Jena D. Hwang and 
    Jiangjiang Yang and
    Ronan Le Bras and
    Oyvind Tafjord and
    Chris Wilhelm and
    Luca Soldaini and 
    Noah A. Smith and 
    Yizhong Wang and 
    Pradeep Dasigi and 
    Hannaneh Hajishirzi
  },
  year = {2024},
  email = {tulu@allenai.org}
}

@article{yu2023megabyte,
  title={Megabyte: Predicting million-byte sequences with multiscale transformers},
  author={Yu, Lili and Simig, D{\'a}niel and Flaherty, Colin and Aghajanyan, Armen and Zettlemoyer, Luke and Lewis, Mike},
  journal={Advances in Neural Information Processing Systems},
  volume={36},
  pages={78808--78823},
  year={2023}
}

@inproceedings{nawrot2023efficient,
  title={Efficient transformers with dynamic token pooling},
  author={Nawrot, Piotr and Chorowski, Jan and Lancucki, Adrian and Ponti, Edoardo Maria},
  booktitle={Proceedings of the 61st Annual Meeting of the Association for Computational Linguistics (Volume 1: Long Papers)},
  pages={6403--6417},
  year={2023},
  doi={10.18653/v1/2023.acl-long.353}
}

@inproceedings{paperno2016lambada,
  title={The {LAMBADA} dataset: Word prediction requiring a broad discourse context},
  author={Paperno, Denis and Kruszewski, Germ{\'a}n and Lazaridou, Angeliki and Pham, Ngoc Quan and Bernardi, Raffaella and Pezzelle, Sandro and Baroni, Marco and Boleda, Gemma and Fern{\'a}ndez, Raquel},
  booktitle={Proceedings of the 54th Annual Meeting of the Association for Computational Linguistics (Volume 1: Long Papers)},
  pages={1525--1534},
  year={2016}
}

@inproceedings{talmor2019commonsenseqa,
  title={{CommonsenseQA}: A Question Answering Challenge Targeting Commonsense Knowledge},
  author={Talmor, Alon and Herzig, Jonathan and Lourie, Nicholas and Berant, Jonathan},
  booktitle={Proceedings of the 2019 Conference of the North American Chapter of the Association for Computational Linguistics: Human Language Technologies, Volume 1 (Long and Short Papers)},
  pages={4149--4158},
  year={2019}
}

@inproceedings{sap2019socialiqa,
  title={Social {IQa}: Commonsense Reasoning about Social Interactions},
  author={Sap, Maarten and Rashkin, Hannah and Chen, Derek and Le Bras, Ronan and Choi, Yejin},
  booktitle={Proceedings of the 2019 Conference on Empirical Methods in Natural Language Processing and the 9th International Joint Conference on Natural Language Processing (EMNLP-IJCNLP)},
  pages={4463--4473},
  year={2019}
}

@article{slagle2024spacebyte,
  title={Spacebyte: Towards deleting tokenization from large language modeling},
  author={Slagle, Kevin},
  journal={Advances in Neural Information Processing Systems},
  volume={37},
  pages={124925--124950},
  year={2024}
}

@inproceedings{nawrot2022hierarchical,
  title={Hierarchical transformers are more efficient language models},
  author={Nawrot, Piotr and Tworkowski, Szymon and Tyrolski, Micha{\l} and Kaiser, {\L}ukasz and Wu, Yuhuai and Szegedy, Christian and Michalewski, Henryk},
  booktitle={Findings of the Association for Computational Linguistics: NAACL 2022},
  pages={1559--1571},
  year={2022}
}

@article{ho2024block,
  title={Block transformer: Global-to-local language modeling for fast inference},
  author={Ho, Namgyu and Bae, Sangmin and Kim, Taehyeon and Jo, Hyunjik and Kim, Yireun and Schuster, Tal and Fisch, Adam and Thorne, James and Yun, Se-Young},
  journal={Advances in Neural Information Processing Systems},
  volume={37},
  pages={48740--48783},
  year={2024}
}

@article{egli2025multiscale,
  title={Multiscale Byte Language Models--A Hierarchical Architecture for Causal Million-Length Sequence Modeling},
  author={Egli, Eric and Manica, Matteo and Born, Jannis},
  journal={arXiv preprint arXiv:2502.14553},
  year={2025}
}

@article{olmo20242,
  title={2 OLMo 2 Furious},
  author={OLMo, Team and Walsh, Pete and Soldaini, Luca and Groeneveld, Dirk and Lo, Kyle and Arora, Shane and Bhagia, Akshita and Gu, Yuling and Huang, Shengyi and Jordan, Matt and others},
  journal={arXiv preprint arXiv:2501.00656},
  year={2024}
}

@article{olmo2025olmo,
  title={Olmo 3},
  author={Olmo, Team and Ettinger, Allyson and Bertsch, Amanda and Kuehl, Bailey and Graham, David and Heineman, David and Groeneveld, Dirk and Brahman, Faeze and Timbers, Finbarr and Ivison, Hamish and others},
  journal={arXiv preprint arXiv:2512.13961},
  year={2025}
}

@article{cobbe2021training,
  title={Training verifiers to solve math word problems},
  author={Cobbe, Karl and Kosaraju, Vineet and Bavarian, Mohammad and Chen, Mark and Jun, Heewoo and Kaiser, Lukasz and Plappert, Matthias and Tworek, Jerry and Hilton, Jacob and Nakano, Reiichiro and others},
  journal={arXiv preprint arXiv:2110.14168},
  year={2021}
}

@misc{yang2025qwen3technicalreport,
      title={Qwen3 Technical Report}, 
      author={An Yang and Anfeng Li and Baosong Yang and Beichen Zhang and Binyuan Hui and Bo Zheng and Bowen Yu and Chang Gao and Chengen Huang and Chenxu Lv and Chujie Zheng and Dayiheng Liu and Fan Zhou and Fei Huang and Feng Hu and Hao Ge and Haoran Wei and Huan Lin and Jialong Tang and Jian Yang and Jianhong Tu and Jianwei Zhang and Jianxin Yang and Jiaxi Yang and Jing Zhou and Jingren Zhou and Junyang Lin and Kai Dang and Keqin Bao and Kexin Yang and Le Yu and Lianghao Deng and Mei Li and Mingfeng Xue and Mingze Li and Pei Zhang and Peng Wang and Qin Zhu and Rui Men and Ruize Gao and Shixuan Liu and Shuang Luo and Tianhao Li and Tianyi Tang and Wenbiao Yin and Xingzhang Ren and Xinyu Wang and Xinyu Zhang and Xuancheng Ren and Yang Fan and Yang Su and Yichang Zhang and Yinger Zhang and Yu Wan and Yuqiong Liu and Zekun Wang and Zeyu Cui and Zhenru Zhang and Zhipeng Zhou and Zihan Qiu},
      year={2025},
      eprint={2505.09388},
      archivePrefix={arXiv},
      primaryClass={cs.CL},
      url={https://arxiv.org/abs/2505.09388}, 
}

@misc{eval-harness,
  author       = {Gao, Leo and Tow, Jonathan and Abbasi, Baber and Biderman, Stella and Black, Sid and DiPofi, Anthony and Foster, Charles and Golding, Laurence and Hsu, Jeffrey and Le Noac'h, Alain and Li, Haonan and McDonell, Kyle and Muennighoff, Niklas and Ociepa, Chris and Phang, Jason and Reynolds, Laria and Schoelkopf, Hailey and Skowron, Aviya and Sutawika, Lintang and Tang, Eric and Thite, Anish and Wang, Ben and Wang, Kevin and Zou, Andy},
  title        = {The Language Model Evaluation Harness},
  month        = 07,
  year         = 2024,
  publisher    = {Zenodo},
  version      = {v0.4.3},
  doi          = {10.5281/zenodo.12608602},
  url          = {https://zenodo.org/records/12608602}
}

\newpage

\appendix

\section{In-depth Analysis of Byte-level Robustness and Boundary Learning}

In this section, we provide a comprehensive evaluation of the robustness of the distilled H-Net and investigate the behavioral patterns of its boundary predictor. 

\subsection{Intrinsic Byte-level Resilience}

Traditional LLMs depend on subword tokenization, which is inherently vulnerable to surface-level disturbances; even slight character-level noise can significantly change token IDs, resulting in semantic collapse. In contrast, byte-level modeling processes raw sequences, theoretically providing greater granularity and robustness. While our distillation framework has been demonstrated to effectively maintain the teacher model's semantic performance on standard benchmarks (as discussed in Section~\ref{sec:experiments}), it is crucial to assess whether the distilled H-Net genuinely inherits the distinctive resilience characteristics of byte-level architectures.

Following the methodology proposed in BLT~\citep{pagnoni2024bytelatenttransformerpatches}, we assess our models using the HellaSwag benchmark, focusing on five types of character-level perturbations: \textit{AntSpeak}, \textit{Drop}, \textit{Random Case}, \textit{Repeat}, and \textit{Uppercase}. Furthermore, we add a \textit{Robustness Score} proposed by H-Net~\citep{hwang2025dynamicchunkingendtoendhierarchical} as a metric, which is calculated as follows:

\begin{equation}
\text{robustness score} = 100 \cdot \left( \frac{\text{perturbed accuracy} - 0.25}{\max(\text{unperturbed accuracy} - 0.25, 0)} \right)
\label{eq:robustness}
\end{equation}

As shown in Table~\ref{tab:robustness}, our distilled Stage 1 models demonstrate a significant level of intrinsic resilience. For Llama-3.2 3B, while the teacher model baseline performs well on clean text, the Stage1 model achieves higher accuracy on several noise types, such as \textit{AntSpeak} (48.78 vs. 46.62) and \textit{Drop} (52.30 vs. 52.03). This indicates that the distillation process does more than merely replicate the teacher model's probability distribution; it enables the H-Net to utilize its byte-to-byte architecture to bypass the fragility of the subword boundaries. Interestingly, the Stage2 models exhibit a decrease in robustness compared to Stage1, particularly in benchmarks such as AntSpeak. This suggests that while further fine-tuning maintains the overall performance, it may introduce a degree of boundary fragility, making the model more sensitive to certain structural perturbations.

\begin{table}[h]
\centering
\caption{\textbf{Robustness Evaluation on HellaSwag.} We reported zero-shot accuracy across the five perturbation types. \textbf{Robustness Score} is computed from clean and noisy accuracy with Eq.~\eqref{eq:robustness} (higher is better). \textbf{Bold} indicates the best performance within each model family.}
\label{tab:robustness}
\small
\setlength{\tabcolsep}{4pt}
\renewcommand{\arraystretch}{0.95}
\begin{tabular}{l|ccc|ccc}
\toprule
\multirow{2}{*}{\textbf{Metric / Model}} & \multicolumn{3}{c|}{\textbf{Qwen-3 4B}} & \multicolumn{3}{c}{\textbf{Llama-3.2 3B}} \\
 & Baseline & Stage1 & Stage2 & Baseline & Stage1 & Stage2 \\
\midrule
\textbf{HellaSwag Original} & 73.64 & 72.47 & 69.70 & 73.76 & 72.90 & 71.20 \\
\textbf{HellaSwag Noise Avg}& 58.22 & 56.25 & 48.86 & 54.10 & 53.29 & 50.60 \\
\midrule
\rowcolor{gray!10} \textbf{Robustness Score} & 68.29 & 65.83 & 53.38 & 59.68 & 59.06 & 55.41 \\
\midrule
\multicolumn{7}{l}{\textit{Breakdown by Perturbation Type}} \\
\midrule
AntSpeak    & \textbf{59.86} & 57.02 & 45.48 & 46.62 & \textbf{48.78} & 48.57 \\
Drop        & 52.94 & \textbf{55.03} & 46.42 & 52.03 & \textbf{52.30} & 48.71 \\
Random Case & \textbf{59.27} & 54.55 & 48.89 & \textbf{53.76} & 51.77 & 49.61 \\
Repeat      & 50.27 & \textbf{50.36} & 42.44 & \textbf{49.05} & 48.74 & 43.17 \\
Uppercase   & \textbf{68.80} & 64.26 & 61.08 & \textbf{69.12} & 64.83 & 62.93 \\
\bottomrule
\end{tabular}
\end{table}

\subsection{Boundary Analysis and Optimization Strategies}
\label{subsec:boundary_optimization}

Despite this intrinsic resilience, the Stage 1 students do not exceed the teacher models in overall robustness. Relative to the teacher, the robustness score decreases from 68.29 to 65.83 for Qwen-3 4B and is nearly unchanged for Llama-3.2 3B (59.68 vs.\ 59.06). To investigate whether the learned boundary predictor limits robustness, we conducted a qualitative analysis of the segmentation behavior of the model under various input conditions.

\begin{figure}[t]
    \centering
    \includegraphics[width=\linewidth]{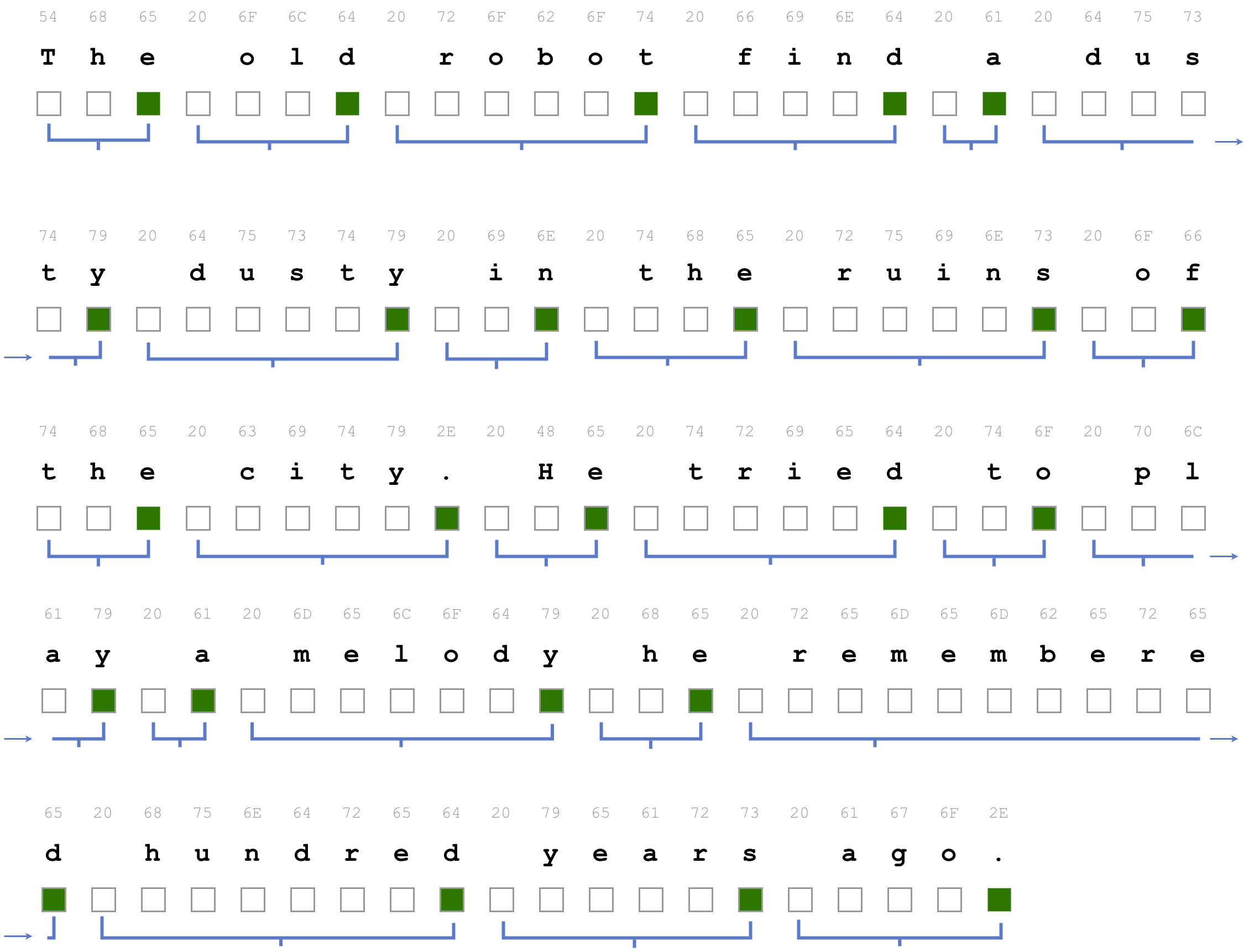}
    \caption{\textbf{Visualization of Learned Token Boundaries.} Chunk boundaries predicted by the learned Routing Module on an example sentence. Each character is shown with its byte value (hex) above it. Filled green squares mark the bytes predicted as chunk ends ($b = 1$), and blue brackets group the bytes of each resulting chunk. The predicted boundaries consistently fall on the byte immediately before a whitespace, which illustrates the \textit{Space Bias} discussed in Appendix~\ref{subsec:boundary_optimization}.}
    \label{fig:router_viz}
\end{figure}

\textbf{Visualization of Boundary Bias.} By visualizing the predicted boundaries (see Figure~\ref{fig:router_viz}), we discovered that the model's segmentation is heavily overfitted to a specific surface pattern: it consistently predicts boundaries at the character immediately preceding a whitespace. This \textit{Space Bias} suggests that the boundary predictor prioritizes shallow lexical cues over deep semantic segmentation. While this heuristic works for clean text, it becomes liable in the presence of perturbations. For instance, in \textit{AntSpeak}, the proliferation of spaces forces the model to segment at every character, whereas in \textit{Drop}, the removal of spaces leaves the model unable to identify logical segments, thereby disrupting the downstream transformer’s processing.

\textbf{Proposed Mitigation Strategies.} To alleviate this bias and force the model to learn more robust semantic-oriented segmentation, we propose two strategies.
\begin{enumerate}
    \item \textbf{Trim Data}: We fine-tune the model on a modified dataset where all whitespaces are removed, forcing the model to rely on character sequences alone to determine boundaries.
    \item \textbf{Whitespace Penalty}: We introduce a penalty term in the boundary loss function to discourage the model from triggering boundaries at whitespace positions, thereby encouraging the exploration of alternative semantic anchor points.
\end{enumerate}

\textbf{Results Analysis.} The results in Table~\ref{tab:optimization_results} reveal a clear trade-off between clean-text accuracy and noise invariance. The \textbf{Trim Data} strategy slightly improves the robustness score over the standard Stage 1 model (59.39 vs.\ 59.06) while largely preserving clean-text accuracy (72.57 vs.\ 72.90). Remarkably, the \textbf{Whitespace Penalty} achieves the lowest sensitivity to noise ($\Delta = 14.45$), effectively creating a "noise-invariant" boundary predictor. However, this stability comes at the cost of a significant drop in \textit{Original} performance (58.50), indicating that while penalizing space boundaries enhances resilience, it may also prevent the model from identifying the standard linguistic boundaries necessary for clean-text reasoning. These findings highlight the challenge of balancing the structural adherence and robustness in byte-level LLMs.

\begin{table}[t]
\centering
\caption{Evaluation of boundary optimization strategies on Llama-3.2 3B. We compare our proposed \textbf{Trim Data} and \textbf{Whitespace Penalty} against the teacher model baseline and the standard boundary learning stage. \textbf{Delta} ($\Delta$) indicates the performance gap between original and noisy benchmarks.}
\label{tab:optimization_results}
\small
\setlength{\tabcolsep}{4pt}
\renewcommand{\arraystretch}{0.95}
\begin{tabular}{l|c|c|cc}
\toprule
\textbf{Metric / Model} & \textbf{Llama} & \textbf{Llama Distill} & \textbf{Llama} & \textbf{Llama Space} \\
& (Teacher Model) & \textbf{Stage 1} & \textbf{Trim Data} & \textbf{Penalty} \\
\midrule
HellaSwag Original & 73.76 & 72.90 & 72.57 & 58.50 \\
HellaSwag Noise Avg & 54.10 & 53.29 & 53.25 & 44.05 \\
\midrule
\rowcolor{gray!10} \textbf{Robustness Score} & 59.68 & 59.06 & 59.39 & 56.85 \\
\textbf{Delta ($\Delta$)} & 19.66 & 19.61 & 19.32 & \textbf{14.45} \\
\midrule
AntSpeak & 46.62 & 48.78 & 48.76 & 39.26 \\
Drop & 52.03 & 52.30 & 54.03 & 44.19 \\
Random Case & 53.76 & 51.77 & 51.59 & 43.39 \\
Repeat & 49.05 & 48.74 & 48.19 & 41.29 \\
Uppercase & 69.12 & 64.83 & 63.70 & 52.10 \\
\bottomrule
\end{tabular}
\end{table}

\subsection{Finetuning on Perturbed Benchmarks: A Limit Study}
\label{subsec:finetune_robustness}

To further investigate the performance limits and adaptation capabilities of different architectures, we conducted a supervised fine-tuning (SFT) experiment. We compare the subword-level Llama (Token) against two of our distilled byte-level H-Net models by fine-tuning all three on the perturbed HellaSwag training set: \textit{Distill}, the Stage 1 student that still generates in the token space, and \textit{Dechunk}, the student after only the Head Adaptation step of Stage 2 (i.e., Stage 2 without end-to-end fine-tuning), which generates entirely in the byte space.

\textbf{Experimental Setup.} All three models are fine-tuned for 10 epochs with a learning rate of $2 \times 10^{-5}$ and a weight decay of $0.1$. This setup aims to assess whether the subword model can "learn" to be robust through data augmentation, and how the byte-level model compares in this intensive adaptation scenario. The results are presented in Table~\ref{tab:finetune_results}.

\textbf{Analysis.} The experimental results highlight the fundamental differences in the adaptability of the two architectures. Notably, the byte-level \textit{Llama distill} model exhibited a significantly higher adaptation ceiling than its subword counterpart. After fine-tuning, it achieved the highest overall Robustness Score (69.93) and consistently surpassed the teacher model across the averaged noise metrics. Critical divergence is evident in the \textit{AntSpeak} task, which serves as a litmus test for tokenizer rigidity. While the subword model's performance remains stagnant at 46.59, despite ten epochs of specific training—underscoring a "hard bottleneck" imposed by fixed vocabulary constraints—our byte-level model successfully adapts to the character-spaced inputs, increasing from 48.78 to 56.07. Furthermore, the byte-level architecture demonstrates superior stability to catastrophic forgetting. Whereas the subword model experiences a decline in clean-text performance (dropping from 73.76 to 71.90) as it struggles to accommodate noisy patterns, the distilled model maintains its original capabilities (73.58), suggesting that its granular representation space allows the integration of noisy features without overwriting the structural knowledge required for standard reasoning.

\begin{table}[t]
\centering
\caption{Comparison of the subword-level model (\textit{Token}) and byte-level models after fine-tuning on perturbed HellaSwag, all based on Llama-3.2 3B. \textit{Distill} is the Stage 1 student, and \textit{Dechunk} is the student after only the Head Adaptation step of Stage 2 (without end-to-end fine-tuning). \textit{Origin} denotes zero-shot performance without noise-specific tuning, while \textit{Finetune} denotes performance after 10 epochs of training on noisy data.}
\label{tab:finetune_results}
\small
\setlength{\tabcolsep}{4pt}
\renewcommand{\arraystretch}{0.95}
\begin{tabular}{l|ccc|ccc}
\toprule
\multirow{2}{*}{\textbf{Metric}} & \multicolumn{3}{c|}{\textbf{Origin}} & \multicolumn{3}{c}{\textbf{Finetuned}} \\
 & Token & Distill & Dechunk & Token & \textbf{Distill} & Dechunk \\
\midrule
HellaSwag Original & \textbf{73.76} & 72.90 & 71.38 & 71.90 & \textbf{73.58} & 70.90 \\
HellaSwag Noise Avg & \textbf{54.10} & 53.29 & 47.47 & 57.23 & \textbf{58.97} & 56.67 \\
\midrule
\rowcolor{gray!10} \textbf{Robustness Score} & \textbf{59.68} & 59.06 & 48.46 & 68.72 & \textbf{69.93} & 69.00 \\
\textbf{Delta ($\Delta$)} & 19.66 & \textbf{19.61} & 23.91 & 14.67 & 14.61 & \textbf{14.23} \\
\midrule
AntSpeak & 46.62 & \textbf{48.78} & 32.32 & 46.59 & \textbf{56.07} & 55.64 \\
Drop & 52.03 & \textbf{52.30} & 49.20 & \textbf{59.32} & 58.30 & 51.74 \\
Random Case & \textbf{53.76} & 51.77 & 48.06 & 57.61 & 56.40 & \textbf{60.70} \\
Repeat & \textbf{49.05} & 48.74 & 44.35 & \textbf{58.55} & 56.38 & 50.57 \\
Uppercase & \textbf{69.12} & 64.83 & 63.44 & 64.10 & \textbf{67.68} & 64.71 \\
\bottomrule
\end{tabular}
\end{table}

\section{End-to-End vs. Sequential Stage-1 Training}
\label{appendix:e2e_vs_sequential}

A central design choice of our framework is to decompose Stage~1 into three sequential steps (embedding alignment $\rightarrow$ joint distillation $\rightarrow$ boundary learning) rather than jointly optimizing all three objectives end-to-end. To empirically validate this choice, we conduct a controlled comparison on OLMo~2 1B against an end-to-end (E2E) baseline that jointly optimizes all three Stage~1 objectives.

\textbf{Setup.} Both runs share the same H-Net architecture (Table~\ref{tab:architecture}), optimizer, learning rate, and batch size. To give the E2E baseline every advantage on the cross-vocabulary distillation objective, we adopt GOLD-style sequence alignment~\citep{patino2025_unlocking_on_policy_distillation_for_any_model_family}, which merges student and teacher distributions at aligned boundary positions before computing KL divergence. This is the state-of-the-art technique for handling boundary and vocabulary mismatch between byte-level students and token-level teachers, ensuring the comparison is not unfairly biased against the joint formulation.

\textbf{Training budget.} Our sequential pipeline processes 30B bytes total ($3\times10$B), but per-step compute is highly asymmetric across the three steps. Only the distillation step performs full-model forward and backward passes (10B bytes, all parameters trained). The embedding alignment step trains only the byte encoder ($<10\%$ of model FLOPs per step), and the boundary learning step trains only the routing head ($<5\%$ of model FLOPs). The sequential FLOP budget is therefore dominated by the 10B distillation step. We train the E2E baseline for 15B bytes with full forward and backward at every step, substantially exceeding the sequential FLOP equivalent. This setup ensures the comparison is not bottlenecked by the compute budget allotted to the E2E baseline.

\textbf{Results.} Table~\ref{tab:e2e_vs_sequential} reports 5-shot accuracy on the same seven benchmarks used in Table~\ref{tab:bolmo_comparison}. The E2E baseline underperforms the sequential pipeline on every task, with an average drop of $7.3$ points. The degradation is most pronounced on benchmarks that depend on stable sequence-level semantic representations: HellaSwag ($-21.6$), PIQA ($-9.3$), and ARC ($-8.9$). Surface-level pattern-matching tasks such as WinoGrande ($-1.9$) and SocialIQA ($-1.1$) are comparatively preserved.

\begin{table}[t]
\centering
\caption{Comparison of sequential and end-to-end (E2E) Stage~1 training on OLMo~2 1B. Both runs use identical optimizer, learning rate, and batch size; the E2E baseline additionally adopts GOLD-style sequence alignment~\citep{patino2025_unlocking_on_policy_distillation_for_any_model_family} and is trained for 15B bytes, exceeding the sequential FLOP equivalent. All numbers are 5-shot accuracy and follow the protocol of Table~\ref{tab:bolmo_comparison}.}
\label{tab:e2e_vs_sequential}
\small
\setlength{\tabcolsep}{8pt}
\renewcommand{\arraystretch}{0.95}
\begin{tabular}{lccc}
\toprule
\textbf{Benchmark} & \textbf{Sequential} & \textbf{E2E} & \textbf{$\Delta$} \\
\midrule
HellaSwag   & 65.4 & 43.8 & $-21.6$ \\
PIQA        & 75.7 & 66.4 & $-9.3$ \\
ARC         & 60.0 & 51.1 & $-8.9$ \\
CSQA        & 65.8 & 60.9 & $-4.9$ \\
MMLU        & 37.6 & 34.4 & $-3.2$ \\
WinoGrande  & 63.9 & 62.0 & $-1.9$ \\
SocialIQA   & 53.8 & 52.7 & $-1.1$ \\
\midrule
\rowcolor{gray!10} \textbf{Average} & \textbf{60.3} & \textbf{53.0} & \textbf{$-7.3$} \\
\bottomrule
\end{tabular}
\end{table}

\textbf{Analysis.} The selective collapse pattern is informative. HellaSwag, PIQA, and ARC require multi-sentence reasoning over long contexts, where stable byte-to-token alignment is essential for the core transformer to function. Joint optimization destabilizes this alignment because the boundary loss and distillation loss compete for the encoder's representational capacity: gradients from boundary learning encourage the encoder to favor segmentation-friendly features, while the distillation gradient simultaneously pulls these representations toward the teacher's token embedding space. Without first anchoring the encoder via embedding alignment---whose necessity is documented in Table~\ref{tab:ablation_stage1}, where the LMB score drops from 70.7 to 31.2 when the alignment step is skipped---these competing gradients prevent the encoder from converging to a stable representation. Notably, this failure persists even with GOLD-style alignment, a method designed specifically to handle cross-vocabulary boundary mismatches, indicating that the optimization difficulty is fundamental rather than a consequence of naive loss combination. These results provide concrete empirical support for our staged design.

\section{Decoupled vs. Integrated Routing Module}
\label{appendix:decoupled_routing}

In the proposed H-Net design, the routing module is seamlessly integrated within the encoder to utilize hidden representations to predict the token boundaries. However, an intriguing alternative is to consider whether the routing logic can be separated from the main semantic encoder to operate as an independent component. To explore this, we developed a decoupled routing module comprising a 4-layer Mamba2 block with a hidden dimension of 512, followed by a linear layer and sigmoid activation head for direct binary boundary classification from raw byte sequences. This standalone router was trained using the same dataset and hyperparameters as in our Stage 1 distillation process.

The experimental results, as outlined in Table~\ref{tab:split_routing}, highlight a notable performance gap between the integrated and decoupled approaches. Most strikingly, the LAMBADA score for the split routing module falls to 39.1, marking a significant decline from the 69.5 achieved by our integrated Stage 1 model. This downward trend is evident across all downstream benchmarks, including HellaSwag, PIQA, and ARC, where the decoupled model lagged consistently.

The observed decline in performance suggests that token boundaries in natural language are not merely local morphological features that can be captured using a shallow, independent module. Rather, the detection of these boundaries is heavily dependent on high-level semantic and syntactic contexts that only a fully developed encoder can provide. When the routing module is compelled to operate in isolation, the system loses the benefit of "feature sharing" between representation learning and structural segmentation. Consequently, even minor inaccuracies in the standalone router's predictions can result in significant chunking errors, leading to the subsequent encoder receiving fragmented byte segments that are no longer aligned with the teacher model's original semantic space. These findings support our architectural decision to integrate the routing mechanism within the hierarchical encoder, because the synergy between semantic understanding and boundary discovery is essential for maintaining the integrity of knowledge transfer in byte-level modeling.

\begin{table}[t]
\centering
\caption{Comparison of performance between the integrated routing module (Stage 1) and a decoupled (Split) routing module. The baseline is the teacher model, Llama 3.2 3B.}
\label{tab:split_routing}
\small
\setlength{\tabcolsep}{4pt}
\renewcommand{\arraystretch}{0.95}
\begin{tabular}{lccccccc}
\toprule
\textbf{Method} & \textbf{LMB} & \textbf{HellA} & \textbf{PIQA} & \textbf{ARC-E} & \textbf{ARC-C} & \textbf{WINO} & \textbf{Open} \\ 
\midrule
Baseline (Teacher) & 70.1 & 73.7 & 76.8 & 74.6 & 46.0 & 67.4 & 41.4 \\
Integrated & 69.5 & 72.9 & 76.1 & 73.6 & 43.9 & 67.2 & 41.1 \\
\midrule
Split Routing & 39.1 & 68.2 & 75.3 & 70.4 & 39.9 & 62.8 & 36.4 \\ 
\bottomrule
\end{tabular}
\end{table}

\section{Teacher Model Scale during Stage 1 Alignment}
\label{appendix:teacher_scale}

In the first stage of our curriculum, we focus on aligning byte-level representations with the latent space of the teacher model. To determine whether the scale of the teacher model plays a decisive role in this process, we evaluated three distinct student-teacher configurations: Llama 3.2 1B distilled from a 3B teacher model and Qwen3 0.6B and 4B distilled from their 4B and 8B versions, respectively. The objective was to observe whether the increased capacity of a larger teacher model translates into more robust representational guidance for byte-level student models.

The comparative results summarized in Table~\ref{tab:teacher_scale_results} reveal that employing a larger teacher model does not offer a distinct advantage. The larger-teacher variants generally underperform the same-scale setup: they score lower on LAMBADA, HellaSwag, and MMLU in all three configurations, although they score higher on ARC-E in all three configurations and on ARC-C for Qwen3 4B. Within the Llama and Qwen families, the standard distillation setup, in which the teacher model is of a similar scale, therefore generally produces better outcomes than larger teacher model variants.

Several factors may explain why larger teacher models do not necessarily enhance distillation outcomes in this context. Notably, contemporary small-scale models, such as Llama 3.2 and Qwen3, often result from the extensive distillation of much larger "frontier" models. This suggests that their latent spaces are highly optimized and information-dense. When a student model is distilled from a slightly larger model within the same family, the potential for "knowledge gain" is limited, while the risk of introducing subtle representational noise or latent misalignments increases. Such discrepancies can complicate student models' learning of byte-to-token mapping, ultimately affecting the overall modeling quality. These findings support our architectural decision to prioritize scale consistency between student models and teacher models as it ensures a more seamless transfer of semantic knowledge during the critical alignment phase.

\begin{table}[t]
\centering
\caption{Comparison of performance across different student-teacher scale configurations. "Baseline" refers to the original subword model, "Stage1" denotes the Stage 1 student using ground-truth token boundaries from the teacher tokenizer, and "Stage1 (Teacher X)" denotes the same setting distilled from a larger teacher model X.}
\label{tab:teacher_scale_results}
\footnotesize
\setlength{\tabcolsep}{3pt}
\renewcommand{\arraystretch}{0.95}
\begin{tabular}{llccccccccc}
\toprule
\textbf{Student Model} & \textbf{Method} & \textbf{LMB} & \textbf{HellA} & \textbf{PIQA} & \textbf{ARC-E} & \textbf{ARC-C} & \textbf{WINO} & \textbf{Open} & \textbf{MMLU} & \textbf{GSM8K} \\ 
\midrule
\multirow{3}{*}{Llama 3.2 1B} & Baseline & 62.2 & 63.8 & 74.2 & 65.3 & 36.0 & 60.0 & 36.6 & 31.0 & 5.3 \\
 & Stage1 & 60.1 & 63.1 & 75.2 & 66.4 & 35.8 & 60.7 & 35.6 & 30.5 & - \\
 & Stage1 (Teacher 3B) & 56.9 & 61.9 & 75.1 & 66.8 & 34.9 & 59.8 & 35.2 & 30.1 & - \\ 
\midrule
\multirow{3}{*}{Qwen3 4B} & Baseline & 69.1 & 73.6 & 77.7 & 78.9 & 51.3 & 70.4 & 40.6 & 73.0 & 84.2 \\
 & Stage1 & 69.7 & 73.6 & 77.8 & 78.1 & 50.9 & 70.0 & 40.4 & 71.9 & 79.6 \\
 & Stage1 (Teacher 8B) & 67.1 & 72.0 & 77.9 & 80.0 & 54.8 & 68.8 & 40.4 & 71.4 & 77.4 \\ 
\midrule
\multirow{3}{*}{Qwen3 0.6B} & Baseline & 53.7 & 53.8 & 69.6 & 65.6 & 38.6 & 59.0 & 34.6 & 52.5 & 46.0 \\
 & Stage1 & 53.5 & 52.8 & 70.1 & 67.3 & 38.1 & 58.3 & 35.6 & 51.1 & 44.4 \\
 & Stage1 (Teacher 4B) & 50.1 & 51.3 & 69.6 & 69.0 & 37.7 & 58.4 & 33.6 & 50.6 & 43.9 \\
\bottomrule
\end{tabular}
\end{table}

\section{Detailed Comparison with Bolmo and Other Baselines}
\label{app:bolmo_comparison}

In this section, we extend our evaluation to include Bolmo 1B alongside other baselines such as BLT 1B and H-Net XL. In this comparison, our Stage 1 and Stage 2 models are distilled from OLMo~2 1B, and all numbers are 5-shot accuracy. The baseline results are taken from Table~2 of \citet{minixhofer2025bolmobyteifyinggenerationlanguage}. Because that table reports averages over a larger evaluation suite, we compute \textbf{Avg} ourselves as the unweighted mean over the seven tasks. As shown in Table \ref{tab:bolmo_comparison}, Bolmo 1B demonstrates strong capabilities, achieving an average accuracy of 60.4\% across the 7 downstream tasks. When comparing our fully byte-level model (Ours Stage 2) against this baseline, we observe a performance gap, with our model attaining an average score of 58.1\%. Specifically, Bolmo 1B outperforms our Stage 2 model on reasoning-heavy benchmarks such as ARC (59.0\% vs. 55.5\%) and HellaSwag (67.0\% vs. 62.5\%). However, our distilled model remains competitive on knowledge-intensive tasks, slightly surpassing Bolmo on MMLU (37.6\% vs. 37.2\%). We also note that BLT 1B remains the strongest baseline in this evaluation suite with an average score of 62.8\%. These results provide a comprehensive view of the current landscape, highlighting that while our distillation framework effectively constructs a functional byte-level model, there remains room for improvement to match the performance of models trained with different objectives or larger computational budgets.

\begin{table}[t]
\centering
\caption{Performance comparison on downstream tasks (5-shot accuracy). We compare our proposed models (Stage 1 \& Stage 2), distilled from OLMo~2 1B, with Bolmo 1B, H-Net XL variants, BLT 1B, and the sub-word reference OLMo~2 1B. Baseline results are taken from \citet{minixhofer2025bolmobyteifyinggenerationlanguage}. \textbf{Avg} is the unweighted mean over the seven tasks, computed by us. The best results among byte-level models are \textbf{bolded}.}
\label{tab:bolmo_comparison}
\small
\setlength{\tabcolsep}{4pt}
\renewcommand{\arraystretch}{0.95}
\begin{tabular}{lccccccc}
\toprule
\textbf{Task} & \textbf{Ours (Stage 1)} & \textbf{Ours (Stage 2)} & \textbf{Bolmo} & \textbf{H-Net (1-stage)} & \textbf{H-Net (2-stage)} & \textbf{BLT} & \textbf{OLMo 2} \\
\midrule
Avg & 60.3 & 58.1 & 60.4 & 57.6 & 59.3 & \textbf{62.8} & 61.9 \\
\midrule
ARC & 60.0 & 55.5 & 59.0 & 61.8 & \textbf{62.3} & 59.9 & 61.4 \\
MMLU & 37.6 & 37.6 & 37.2 & 37.5 & 38.7 & \textbf{40.6} & 40.4 \\
CSQA & 65.8 & 62.6 & 64.2 & 61.4 & 62.4 & \textbf{69.2} & 66.0 \\
HellaSwag & 65.4 & 62.5 & 67.0 & 60.2 & 63.6 & \textbf{71.0} & 68.9 \\
WinoGrande & 63.9 & 63.5 & 65.7 & 58.9 & 60.9 & \textbf{67.0} & 65.2 \\
SocialIQA & 53.8 & 52.9 & \textbf{54.7} & 50.1 & 52.9 & 54.6 & 55.1 \\
PIQA & 75.7 & 72.3 & 74.9 & 73.6 & 74.0 & \textbf{77.3} & 76.4 \\
\bottomrule
\end{tabular}
\end{table}

\begin{table}[h]
\centering
\caption{Student model architecture details.}
\label{tab:architecture}
\small
\setlength{\tabcolsep}{4pt}
\renewcommand{\arraystretch}{0.95}
\begin{tabular}{lccc}
\toprule
\textbf{} & \textbf{H-Net 3B} & \textbf{H-Net 4B} & \textbf{H-Net 1B} \\
\midrule
\multicolumn{4}{l}{\textbf{Global Model}} \\
\hspace{1em}Initialized from & Llama 3.2 3B & Qwen 3 4B & OLMo 2 1B \\
\midrule
\multicolumn{4}{l}{\textbf{Encoder}} \\
\hspace{1em}Dimension & 1536 & 1536 & 1536 \\
\hspace{1em}Layer Type & Mamba2 + FFN & Mamba2 + FFN & Mamba2 + FFN\\
\hspace{1em}Num. Layers & 4 & 4 & 4 \\
\hspace{1em}Mamba2 \\
\hspace{2em}chunk\_size & 256 & 256 & 256 \\
\hspace{2em}d\_conv & 4 & 4 & 4 \\
\hspace{2em}d\_state & 128 & 128 & 128 \\
\hspace{2em}expand & 2 & 2 & 2 \\
\hspace{1em}FFN \\
\hspace{2em}Expansion Dim. & 8192 & 8192 & 8192 \\
\hspace{2em}Nonlinearity & SwiGLU & SwiGLU & SwiGLU \\
\hspace{2em}Layer norm & RMSNorm & RMSNorm & RMSNorm \\
\midrule
\multicolumn{4}{l}{\textbf{Local Decoder}} \\
\hspace{1em}Dimension & 1536 & 1536 & 1536 \\
\hspace{1em}Layer Type & Mamba2 + FFN & Mamba2 + FFN & Mamba2 + FFN \\
\hspace{1em}Num. Layers & 4 & 4 & 4 \\
\hspace{1em}Mamba2 \\
\hspace{2em}chunk\_size & 256 & 256 & 256 \\
\hspace{2em}d\_conv & 4 & 4 & 4 \\
\hspace{2em}d\_state & 128 & 128 & 128 \\
\hspace{2em}expand & 2 & 2 & 2 \\
\hspace{1em}FFN \\
\hspace{2em}Expansion Dim. & 8192 & 8192 & 8192 \\
\hspace{2em}Nonlinearity & SwiGLU & SwiGLU & SwiGLU \\
\hspace{2em}Layer norm & RMSNorm & RMSNorm & RMSNorm \\
\bottomrule
\end{tabular}
\end{table}

\begin{table}[h]
\centering
\caption{Training details.}
\label{tab:HyperParameters}
\small
\setlength{\tabcolsep}{4pt}
\renewcommand{\arraystretch}{0.95}
\begin{tabular}{lccc}
\toprule
\textbf{} & \textbf{Llama 3.2 3B} & \textbf{Qwen 3 4B} & \textbf{OLMo 2 1B} \\
\midrule
\multicolumn{4}{l}{\textbf{Stage 1}} \\
\hspace{1em}Total Training Bytes & 30B & 30B & 30B \\
\hspace{1em}Batch Size & 256 & 256 & 256\\
\hspace{1em}Seq. Length. (Bytes) & 8192 & 8192 & 8192 \\
\hspace{1em}LR Schedule & Warmup + Cos Decay & Warmup + Cos Decay & Warmup + Cos Decay \\
\hspace{1em}Step 1 \\
\hspace{3em}Peak LR & 1e-3 & 1e-3 & 1e-3 \\
\hspace{3em}Warmup Ratio & 0.2 & 0.2 & 0.2 \\
\hspace{1em}Step 2 \\
\hspace{3em}Peak LR (Encoder) & 1e-3 & 1e-3 & 1e-3 \\
\hspace{3em}Peak LR (Main Transformer) & 2e-5 & 2e-5 & 2e-5 \\
\hspace{3em}Warmup Ratio & 0.01 & 0.01 & 0.01 \\
\hspace{1em}Step 3 \\
\hspace{3em}Peak LR & 1e-3 & 1e-3 & 1e-3 \\
\hspace{3em}Warmup Ratio & 0.1 & 0.1 & 0.1 \\
\hspace{1em}Weight Decay & 0.1 & 0.1 & 0.1 \\
\hspace{1em}Optimizer & AdamW & AdamW & AdamW \\
\hspace{1em}$\beta_1, \beta_2$ & 0.9, 0.95 & 0.9, 0.95 & 0.9, 0.95 \\
\hspace{1em}Max. Grad. Norm & 1.0 & 1.0 & 1.0 \\
\midrule
\multicolumn{4}{l}{\textbf{Stage 2}} \\
\hspace{1em}Total Training Bytes & 95B & 95B & 95B \\
\hspace{1em}Batch Size & 256 & 256 & 256 \\
\hspace{1em}Seq. Length. (Bytes) & 8192 & 8192 & 8192 \\
\hspace{1em}LR Schedule & Warmup + Cos Decay & Warmup + Cos Decay & Warmup + Cos Decay \\
\hspace{1em}Step 1 \\
\hspace{3em}Peak LR (Decoder) & 1e-3 & 1e-3 & 1e-3 \\
\hspace{3em}Warmup Ratio & 0.1 & 0.1 & 0.1 \\
\hspace{1em}Step 2 \\
\hspace{3em}Peak LR (Encoder + Decoder) & 4e-5 & 4e-5 & 4e-5 \\
\hspace{3em}Peak LR (Main Transformer) & 2e-5 & 2e-5 & 2e-5 \\
\hspace{3em}Warmup Ratio & 0.01 & 0.01 & 0.01 \\
\hspace{1em}Weight Decay & 0.1 & 0.1 & 0.1 \\
\hspace{1em}Optimizer & AdamW & AdamW & AdamW \\
\hspace{1em}$\beta_1, \beta_2$ & 0.9, 0.95 & 0.9, 0.95 & 0.9, 0.95 \\
\hspace{1em}Max. Grad. Norm & 1.0 & 1.0 & 1.0 \\
\bottomrule
\end{tabular}
\end{table}

\section{Training Hyperparameters}
\label{appendix:hyperparameters}
Full architecture hyperparameters are shown in Table \ref{tab:architecture}, and full training in Table \ref{tab:HyperParameters}.

\end{document}